\documentclass[11pt,a4paper]{article}
\usepackage[nohyperref]{emnlp2018}
\usepackage{times}
\usepackage{latexsym}
\usepackage[ruled,linesnumbered]{algorithm2e}
\usepackage{algorithmic}

\usepackage{url}

\usepackage{graphicx}
\usepackage{enumitem} 
\usepackage{amsmath}
\usepackage{amssymb}
\usepackage{breqn}
\usepackage{todonotes}
\usepackage{multirow}

\aclfinalcopy % Uncomment this line for the final submission
\newcommand\BibTeX{B{\sc ib}\TeX}
\newcommand\confname{EMNLP 2018}
\newcommand\conforg{SIGDAT}
\newcommand*\samethanks[1][\value{footnote}]{\footnotemark[#1]}

\DeclareMathOperator*{\argmax}{argmax}
\DeclareMathOperator*{\argmin}{argmin}

\title{Free as in Free Word Order: An Energy Based Model for Word Segmentation and Morphological Tagging in Sanskrit}

\author{Amrith Krishna\textsuperscript{1}, Bishal Santra\textsuperscript{1}, Sasi Prasanth Bandaru\textsuperscript{2}\thanks{\ Work done while the authors were at IIT Kharagpur}, Gaurav Sahu\textsuperscript{3}\samethanks \\\textbf{Vishnu Dutt Sharma\textsuperscript{4}\thanks{\ Part of the work was done while the authors were at IIT Kharagpur}, Pavankumar Satuluri\textsuperscript{5}\samethanks \ and Pawan Goyal\textsuperscript{1}} \\
  \textsuperscript{1}Dept. of Computer Science and Engineering, IIT Kharagpur \\
  \textsuperscript{2}Flipkart, India \textsuperscript{3}School of Computer Science, University of Waterloo \\
  \textsuperscript{4}American Express India Pvt Ltd \textsuperscript{5}Chinmaya Vishwavidyapeeth\\
  {\tt amrith@iitkgp.ac.in, bsantraigi@gmail.com,}\\{\tt
  pawang@cse.iitkgp.ernet.in} \\
  }

\date{}

\begin{document}
\maketitle
\begin{abstract}
The configurational information in sentences of a free word order language such as Sanskrit is of limited use. Thus, the context of the entire sentence will be desirable even for basic processing tasks such as word segmentation. We propose a structured prediction framework that  jointly solves the word segmentation and morphological tagging tasks in Sanskrit.  %takes the `free' word order structure of the language into consideration and not a model that is `free' of any structural considerations. %for performing structured prediction tasks in Sanskrit, which focuses on its free word order nature. In this paper, we construct a 
We build an energy based model where we adopt approaches generally employed in graph based parsing techniques \cite{mcdonald2005simple,carreras2007experiments}. Our model outperforms the state of the art with an F-Score of 96.92 (percentage improvement of 7.06\%) while using less than one tenth of the task-specific training data. We find that the use of a graph based approach instead of a traditional lattice-based sequential labelling approach leads to a percentage gain of 12.6\% in F-Score for the segmentation task.% The performance gain 7.06\% in F-Score we obtain justifies the computational overhead incurred in using our approach  %For a low resource language like Sanskrit, this is a desirable property. 
\footnote{The code and the pretrained edge vectors (\S \ref{PCRW}) used in this work are available at \url{https://zenodo.org/record/1035413\#.W35s8hjhUUs}} 
\end{abstract}

\section{Introduction}
\label{intro}

Sanskrit, a morphologically-rich and a free word order language~\cite{Ambakulkarni2015}, poses a series of challenges even for automation of basic processing tasks such as word segmentation.
%The linguistic peculiarities expressed by Sanskrit poses its own challenges in processing them, even for the basic tasks such as word segmentation.
The recent surge in the digitisation efforts for archiving the works ranging from the pre-classical to modern times \cite{dcsOliver} has led to a growing demand for such tools
%Word segmentation is an essential task of identifying meaningful segments from the written scripts of a language. 
  %, a language that used to be the `lingua franca' of the scientific discourse in ancient India, 
 \cite{goyal2012distributed,2006-Huet-2}.   %Such texts are increasingly getting digitised. 
%Facilitating tools for effective processing of such writings is pivotal in ensuring their accessibility.  
We propose a structured prediction approach that jointly solves the word
segmentation and morphological tagging tasks
for Sanskrit. 

\begin{figure*}[!htb]
\centering
\minipage{\textwidth}
  \includegraphics[width=\textwidth]{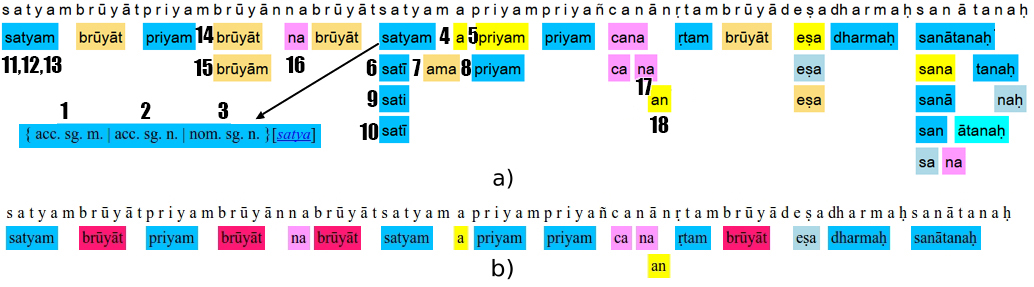}
\vspace{-7mm}
\caption{a) All the phonetically valid segmentations (\href{http://sanskrit.inria.fr/cgi-bin/SKT/sktgraph?lex=SH&st=t&us=f&cp=t&text=satya.mbruuyaatpriya.mbruuyaannabruuyaatsatyamapriya.mpriya.mcanaan.rtambruuyaade.sadharma.hsanaatana.h&t=VH&topic=&mode=g}{link}) for `{\sl satya{\d m}br\=uy\=atpriya{\d m}br\=uy\=anna\-br\=uy\=atsatyamapriya{\d m}priya{\d m}can\=an{\d r}tambr\=uy\=ade{\d s}adharma{\d h}san\=atana{\d h}}' from ({\sl subh\=a{\d s}itam}) as output by Sanskrit Heritage Reader (SHR) and b) correct segmentation selected from the candidate space.}

  \label{herit}
\endminipage
\vspace{-4mm}
\end{figure*}

The computational problems arising from the mechanical treatment of Sanskrit fall somewhere between speech recognition and the analysis of written text \cite{2004-Huet-1}. For instance, consider Figure \ref{herit}a which shows all the phonetically valid word splits for a Sanskrit poetic verse\footnote{A saying from {\sl subh\=a{\d s}itam} text: ‘One should tell the truth,
one should say kind words; one should neither tell harsh truths, nor flattering lies; this is a rule for all times.}.  The written representation in Sanskrit is actually a phonemic stream \cite{2004-Huet-1}. The constructions often undergo phonetic transformations at the juncture of successive words, similar to what one observes in connected speech %\footnote{Usages like wanna, gotta etc. are common examples for connected speech in English} 
\cite{morris2004and,shieber2003comma}%, well known by the term ``\textit{sandhi}''
. These transformations obscure the word boundaries and often modify the phones at these word boundaries. In Sanskrit, these transformations get reflected  in writing as well. This is primarily due to %the oral tradition that dominated the sphere of learning  and 
the presence of an advanced discipline of phonetics in Sanskrit which explicitly described euphonic assimilation as \textit{sandhi} \cite{goyal2016design}. For instance, words prefixed with numbers 14 and 15 in Figure \ref{herit}a are valid candidates in spite of the phonetic differences they posses from that of the original sentence.

Sanskrit is rich with syncretisms \cite{crystal2011dictionary} and homonyms.  For example, the surface form `{\sl sat{\=\i}}', prefixed with numbers 6 and 10,  are homonyms, %with different roots and morphological classes. 
 while the root `\textit{satya}' generates identical surface form for three different morphological classes leading to syncretism (1 to 3 in Figure \ref{herit}a).
%the surface `Satyam'  represents syncretism as three identical surface forms of the root \textit{satya}  but in different morphological classes are represented by the numbers 1,2 and 3.
%identical surface forms for a given root under different morphological classes is prevalent in Sanskrit. 
Hence, in addition to segmentation,  the morphological analysis of the segmented word forms will be %$Similar to other morphologically rich languages, identifying the inflected word forms in Sanskrit, without the morphological analysis  will be of limited use \cite{krishna-satuluri-goyal:2017:LaTeCH-CLfL}. Morphological analysis will be 
critical for reducing the ambiguity in further downstream tasks such as syntactic analysis. The sentence construction in the language follows weak non-projectivity \cite{havelka:2007:ACLMain} permitting the words to have a relatively free word order structure \cite{Ambakulkarni2015}. The language is all the more lenient for poetic constructions~\cite{scharf2015distinctive,Ambakulkarni2015}, where arranging the words to adhere to metrical constraints is a bigger concern \cite{melnad2013meter}.
%The words in the solution shown in Figure \ref{herit}b can be permuted in 13,621,608,000\footnote{Detailed in \S 1 of the supplementary material} ways, without affecting the syntactic and semantic validity of the construction. 
Hence, the whole input context is desirable when making each prediction at the output~\cite{bahdanau2014neural}, even for preliminary tasks such as segmentation in Sanskrit \cite{reddy2018seg}.

%The word boundaries in Sanskrit, similar to Chinese, are not explicit. Additionally,  

The word splits in Figure \ref{herit} are based on the analysis by a lexicon driven analyser, Sanskrit Heritage Reader (SHR)\footnote{Available at \href{http://sanskrit.inria.fr/cgi-bin/SKT/sktgraph?lex=SH&st=t&us=f&cp=t&text=satya.mbruuyaatpriya.mbruuyaannabruuyaatsatyamapriya.mpriya.mcanaan.rtambruuyaade.sadharma.hsanaatana.h&t=VH&topic=&mode=g}{http://sanskrit.inria.fr/}, SHR is a lexicon-driven segmenter which produces all the valid word splits. An interface is provided for manual selection of a solution~\cite{goyal2016design}}. A total of 1,056 combinations can be formed from the word splits, such that each of those combinations is a solution which covers the entire input.
%From the given word splits, we can form a total of 1056 unique segmentation s. 
We call such a solution as an `\textit{exhaustive segmentation'}. Out task is to find an `\textit{exhaustive segmentation'}, which is also semantically valid. Figure \ref{herit}b shows the semantically valid solution for the sentence.

We propose our structured prediction framework as an energy based model \cite{lecun2006tutorial}. Considering the free word-orderness, morphological richness and the phenomena of \textit{Sandhi} in Sanskrit, we adopt a graph based treatment for a given input sentence as shown in Figure \ref{archi}. All the word splits, as predicted by SHR, are treated as the nodes in the graph. Every pair of nodes that can co-occur in at least one `\textit{exhaustive segmentation}'\footnote{For instance, segments 6 and 7 in Figure ~\ref{herit}a are connected, while 6 and 9 are not.} forms directed edges in both the directions. By construction, any subset of nodes that forms a maximal clique will be an `\textit{exhaustive segmentation}'. %, i.e. a solution when linearly arranged do not leave any position unfilled in the input sentence. 
%The semantically valid split, in this case the one shown in Figure \ref{herit}b, is going to be one of the such maximal cliques. 
%The solution is going to be a particular subset of the vertex set and the set of nodes will form a clique .%A subset of the vertex set in the graph forms the solution. 
We formalise our task as the search for a maximal clique.
 %The  propsed model searches for the minimum weighted maximal clique of this graph \cite{mcdonald2005non}. The edges in the graph are featurised and  we assume that the 
%The energy of the clique can be factorised as the sum of the individual energies of the edges \cite{mcdonald2005non,ishikawa2011transformation}. We use a greedy maximal clique selection approach for inference over the graph.
The graph structure eliminates the sequential nature of the input, while the greedy maximal clique selection inference policy of ours can take the entire input context into consideration. We hypothesise that both of these will be beneficial for processing constructions in Sanskrit.  %Inspired from the structured prediction model by \newcite{mcdonald2005non} proposed for dependency parsing, we search for a minimum weighted maximal clique from the graph. 

The major contributions of our work are:

\begin{enumerate}[leftmargin=*,itemsep=0.05em]
\item We propose the first model that performs both word segmentation and morphological tagging for Sanskrit as suggested by \newcite{krishna-satuluri-goyal:2017:LaTeCH-CLfL}; 
the combined task reports an F-Score of 90.45. 
\item We obtain an F-Score of 96.92 for the word segmentation task, an improvement of 7.06\% over the state of the art, a seq2seq model with attention \cite{reddy2018seg}.
\item We achieve the results with less than one-tenth of the training data that \newcite{reddy2018seg} uses, a desirable outcome for a low resource language such as Sanskrit. The pre-training in the form of morphological constraints to form edge vectors enables this. 
\item We propose a scheme that uses the Path Ranking Algorithm \cite{lao2010relational}
to automate the feature selection and the feature vector generation for the edges. This eliminates the need for manual feature engineering. 
%\item We propose a generic framework for Structured prediction tasks in Sanskrit. We show how our framework can be extended to perform other structured prediction tasks as well.
\end{enumerate}

\section{Proposed Architecture}
\label{inGraph}

\begin{figure}[!htb]
\minipage{0.5\textwidth}
\centering

  \includegraphics[width=\textwidth]{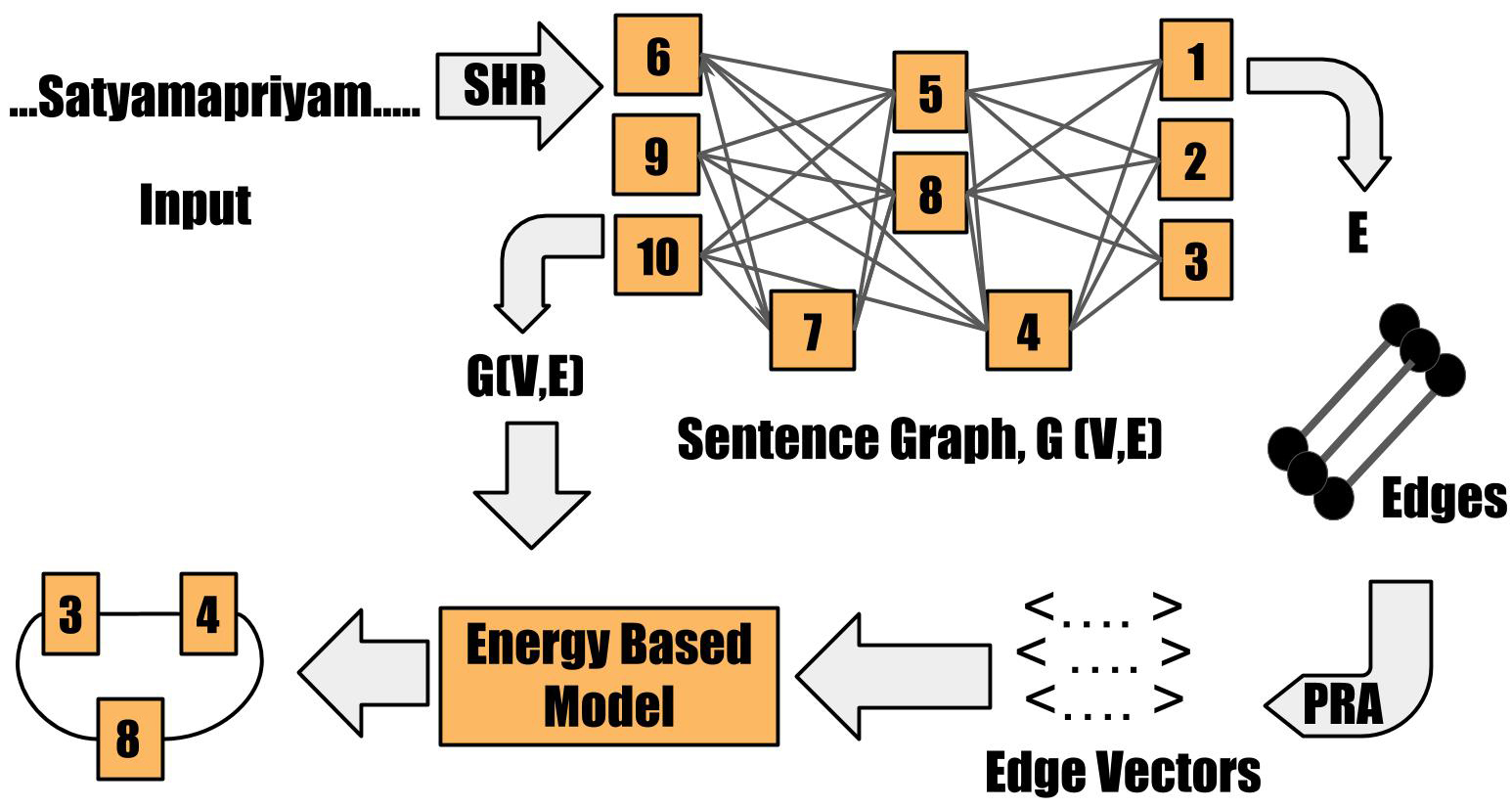}

\caption{Architecture for the proposed model. Word-splits for `\textit{satyamapriyam}', a sub-sequence of the sentence in Figure \ref{herit}a are considered here. The nodes are numbered from 1 to 10 and are marked with the same in Figure \ref{herit}a. For simplicity, we assume that words in nodes 4 to 10 have only one morphological analysis each.}

  \label{archi}
\endminipage
\vspace{-3mm}
\end{figure}

%The design of our system was motivated by the following linguistic peculiarities expressed by sentence constructs in Sanskrit corpora. 

%However, since we need to identify  the minimum cost spanning tree for a sub-graph, this can be reduced to the Steiner tree problem. The Steiner tree problem is an NP-Hard problem, and we use an approximation scheme by employing Prim's algorithm for finding a sub-optimal spanning tree. 

%1) \textit{\textbf{Sandhi}} - %Multiple words in a sentence join together to form a single chunk via phonetic transformations at the word boundaries called as ``{\sl sandhi}''. 
%The phonetic transformation at word boundaries, also known as {\sl sandhi}, is primarily an outcome of the euphonic assimilation in speech, that gets reflected in writing as well \cite{goyal2016design}. The transformations can be substitution of phonemes (Figure \ref{compeq}a), elision of a phoneme (Figure \ref{compeq}b-c), insertion of phonemes (Figure \ref{compeq}d) or a combination of any of these. %Due to sandhi, the words often have overlapping positions in their joint forms (Figure \ref{compeq}b-c).   %The phonetic transformation can take place between any two successive words with compatible phones at their boundaries, as 

\if{}
\begin{figure}[!htb]
\centering
\vspace{-0.2em}
\minipage{\columnwidth}
\includegraphics[width=\columnwidth]{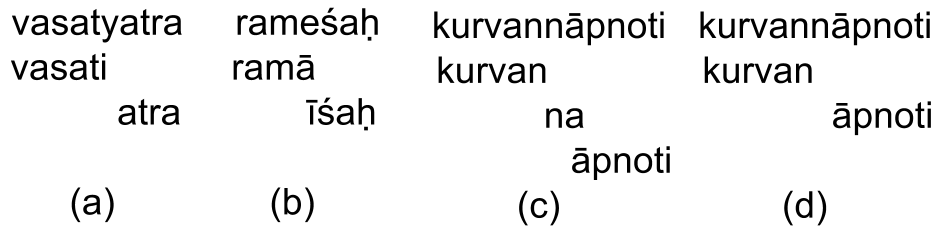}
\vspace{-7mm}
\caption{Instances of sandhi formation in Sanskrit. a) Phonetic transformation of `{\sl i}' and `{\sl a}' to `{\sl ya}'. b) `{\sl \=a}' and `{\sl \=\i}' at the word boundaries join together to form a single `{\sl e}' \protect \cite{goyal2016design}. c) and d) Two possible analyses for the sandhied chunk `{\sl kurvann{\=a}pnoti}' \protect \cite{krishna2016segmentation}, where both are negations of one another.}
\label{compeq}
\vspace{-6mm}
\endminipage\hfill
\end{figure}
The proximity between phonemes is the sole criteria for applying {\sl sandhi}. Unlike in compounds, this does not modify the words syntactically or semantically \cite{krishna2016segmentation}. %Figure \ref{compeq} shows different instances of {\sl sandhi} transformations. 
The rules for Sandhi are well documented in Sanskrit grammatical tradition. 
\fi{}

Given an input construction, we obtain our search space of possible word splits using SHR as shown in Figure \ref{herit}. The search space represents all the possible exhaustive segmentations with possible gaps and overlaps between the word splits in each of the exhaustive segmentation \cite{kudo2006mecab,wordGraph,wolf1977hwim}.\footnote{The word splits in an exhaustive segmentation often overlap and sometimes leave gaps by virtue of Sandhi. For examples, please refer the \S 1 supplementary material.} In such a setting, representing the search space as a lattice \cite{kudo2006mecab,smith-smith-tromble:2005:HLTEMNLP} has been a popular approach for fusional languages \cite{goldberg-tsarfaty:2008:ACLMain,cohen-smith:2007:EMNLP-CoNLL2007,hatori-EtAl:2012:ACL2012}. %Lattice parsing is a well known means of statistical sequence labelling approach for joint modelling of word segmentation, morphological tagging  and is extended to dependency parsing as well \cite{goldberg-tsarfaty:2008:ACLMain,seeker2015graph}. a popular model for handling segmentation and morphological analysis in morphologically rich languages. 
In a lattice, there will be edges only between the adjacent word splits of an exhaustive segmentation. %Explicit start and end markers indicate the order of processing the input.
%The SHR output then needs to be converted to a single graph structure which forms the input to our model. The SHR output  Typically, such an output is modelled as a lattice , where 
We deviate from this norm in a minor yet fundamental way. In the search space, we choose to add edges between every pair of word splits that are part of a single exhaustive segmentation.  Henceforth, we will refer this structure as the sentence graph $G$. We then employ our minimum cost maximal clique finding energy based model on the sentence graph $G$.  Figure \ref{archi} shows the proposed architecture of the model. It shows the sentence graph $G$ for `\textit{satyamapriyam}', a sub-sequence of the sentence in Figure \ref{herit}a.

% The lattice parsing approaches can be seen as a special case of Graph Transformer Networks \cite{lecun1998gradient}  in the energy based model nomenclature \cite{legrand2014recurrent,Collobert:2011:NLP:1953048.2078186}. 

Our current design choice results in a denser graph structure as input and a computationally expensive inference. Such a choice 
%Our choice to go with a dense graph construction scheme and then use a computationally expensive clique selection approach as the inference%, instead of the standard  sequential labelling model on the lattice structure, 
requires justification. Currently, there exist digitised versions of texts which spans over a period of 3000 years categorised into pre-classical literature (1500 BCE - 100 BCE), classical literature (300 CE - 800 CE) and modern literature (900 CE to now). %The stylistic analysis of digitised literature from the aforementioned periods shows high variations \cite{hellwig2008extracting}. 
\newcite{hellwig2008extracting} assert that the assumption that Sanskrit
syntax has remained unchanged over an interval of over 3000 years is not valid.  \newcite{Ambakulkarni2015} notes that the constructions in prose generally follow weak non-projectivity \cite{havelka:2007:ACLMain,maier2009characterizing}. \newcite{Ambakulkarni2015} also observes that constructions in verses violate weak non-projectivity especially with the adjectival and genitive relations. 
A large  number of texts are written in verses or to complicate things further, they are written as a combination of prose and verse. % The verses often tend to violate the restrictions put with respect to free word orderness of the language when writing in prose, as the primary concern there is to satisfy the meter of the verse \cite{Ambakulkarni2015}. %Considering the fact that Sanskrit is a resource scarce language, categorising and sub categorising content based on the time periods and then on the writing styles with further subdivision on the meter styles will aggravate the current issue of data scarcity. 
A lack of consensus among the experts on a common set of rules for works across the different time spans, and the enormous effort in categorising constructions based on their writing styles, motivated us
to use this graph construction scheme which is agnostic to word order. %stems from
%\textbf{TODO: Are these sentences needed? Not very clear. The notion of context windows \cite{sahlgren2006word} for obtaining distributional information \cite{muller-schuetze:2015:NAACL-HLT,clark2003combining} is not feasible here and we take the overhead on the system design. But, for the current corpus we tested, limiting the context window do not affect much on the sparsity. }

%Given a Sanskrit sentence, we obtain all the valid candidate word segments using SHR. The candidate segments so obtained is then converted into a

\paragraph{Graph Formation:\footnote{Our graph construction approach is explained using finite state methods in \S 1 of the supplementary material}}  In Figure \ref{herit}a, identical surface-forms with the same root are grouped together and displayed as a single entity. But we consider, every unique combination of root, morphological class and the word position in SHR as a separate node in $G(V,E)$. Hence the surface-from \textit{satyam}, appears as 6 separate nodes numbered from 1-3 and 11-13 in Figure \ref{herit}a. Here the nodes 1-3 differ from each other in terms of their morphological classes. The nodes 1 and 11 differ only in terms of their position owing to its repeated occurrence in the input. The position information is opaque to our proposed system and is used only in forming the nodes for the sentence graph. %which use the graph as input are not provided with it.  Hence this information is not part of the inference. %the position information of the words is not  anymore exposed to the system and  
During the inference, we consider all the pairwise potentials as contexts for each of the prediction made in the search space.  %The surface form of a word can be deterministically obtained from the lemma and morphological class of a word. In Figure \ref{herit}a,
The edges in our model should capture the likeliness of two nodes to co-occur in the final solution. Hence, every pair of nodes that can co-occur in an `exhaustive segmentation' forms two directed edges, one each at either of the directions.%\footnote{By virtue of \textit{Sandhi}, words appearing in overlapping positions, such as the candidates 17 and 18 in Figure \ref{herit}a,  also can co-exist in an exhaustive segmentation.} % %appearing in overlapping positions and do not follow any of the `\textit{Sandhi}' rules \cite{goyal2016design} will not have edges between them and  is called as a conflicting pair. %However by virtue of , , %i.e., euphonic assimilation of phonemes at the word boundaries, 
 %For example, will form edges between them as there exists a valid {\sl sandhi} rule.%, but 19 and 20 are conflicting.

 %We define $F$ as the set of edge vectors such that there exists an edge vector $f_{ij} \in F$ for each edge  $e_{ij} \in E$. 

%For every edge  there exists a corresponding feature vector representation denoted as $f_{pq} \in F$, where $F$ is the set of all edge vectors. 

%Every unique segment generated by SHR is a node in the graph. Only nodes which are conflicting, i.e. nodes which are suggested by the SHR as alternatives of one another do not have edges between them. %We add edges between every pair of non-conflicting nodes, i.e., between every pair of nodes that can potentially co-exits in the final solution.

\paragraph{Energy Based Model (EBM) Architecture:}
    %\subsection{Energy Function}  
    
    %We construct the input for the model as a graph $G(V,E)$, which we call as sentence graph.  
    
    %of theremaining variables, and higher energies to incorrect values
Our approach is inspired from the graph based parsing approaches employed generally for dependency parsing \cite{mcdonald2005non,carreras2007experiments} and follows a likewise structured prediction paradigm \cite{Taskar:2005:LSP:1102351.1102464}. 
Specifically, we use an EBM where we model our joint task as search for a maximal clique with minimum energy.  Learning consists of finding an energy function that associates lower energies to cliques with increasing similarity to the correct clique. The correct clique configuration will have the lowest energy~\cite{lecun2006tutorial}. %Theloss functional, minimised during learning, measures the quality of the function learnt. 
The inference policy, a maximal clique selection algorithm, is used to  make the predictions.%, i.e., the solution with the minimum energy. 
    
We follow an approach similar to the arc-factored approaches in graphs \cite{mcdonald2005non,carreras2007experiments}, where the total energy of a maximal clique, $T_{i} = (V_{T_i}, E_{T_{i}})$, is decomposed as the summation of energies of its edges~\cite{ishikawa2011transformation}.  
\begin{dmath*}
\begin{small}
        \mathcal{S}(T_i) = \sum_{e_{pq} \in E_{T_{i}}} \mathcal{S}(\vec{e}_{pq})
\end{small}
\end{dmath*}

where, $V_{T_i} \subseteq V, E_{T_{i}} \subseteq E$. The edges are featurised. For an edge $e_{pq} \in E$, the features are represented as a vector, denoted by $\vec{e}_{pq}$. %$En_f(\cdot)$ is a function, 
For a given edge, the energy function, $\mathcal{S}(\cdot) : [0,\infty)^{|\vec{e}|} \rightarrow (-\infty,\infty)$, takes the edge feature vector and produces a scalar value as the energy assignment.

\paragraph{Loss Function and Training:}
We use Hinge Loss \cite{Altun-etAl:2003:EMNLP,taskar2004max} as our loss function.  The hinge loss is formulated such that it increasingly penalises those cliques, sampled by our inference algorithm, with increasingly more number of wrong segmentation candidates. We minimise the hinge loss $L$ which is defined as
\begin{dmath*}
\begin{small}
L=max(0, \mathcal{S}(T_{GT}) - \argmin_{T_i\in A^{Q}}(\mathcal{S}(T_{i}) - \Delta(T_{i}))
\end{small}
\end{dmath*}
\vspace{-2mm}
Here, $A^{Q}$ denotes the set of all the unique maximal cliques and $T_{GT}$ denotes the maximal clique corresponding to the ground truth. 

The margin  $\Delta(T_i)$ is defined as $\Delta(T_{i}) = | V_{T_{i}} - V_{GT} |^2$.
%The loss function ensures that the energy of gold standard maximal clique $T_{GT}$ should always be less than or equal to that of any maximal clique $T_i\in A^{Q}$ by a margin of  where
%Since the aim is to obtain any of the gold standard spanning arborescence, we define our loss function to ensure that the minimum energy of any gold standard spanning arborescence $A^G_m$ should be less than any of the spanning arborescences $A^Q_n$ , as may be output by Prim's algorithm (for different root nodes) by a margin $\Delta(A^Q_n)$. This can be taken care of by comparing the minimum energy gold standard arborescence and minimum energy spanning arborescences (output of Prim's algorithm).
We minimise the given loss function using gradient descent method. The network parameters are updated per sentence using back-propagation. The hinge loss function is not differentiable at the origin. Hence, we use the subgradient method %for the back-propagation 
to update the network parameters~\cite{socher2010learning,ratliff2007online}. We use a multi-layer perceptron network with a single hidden layer and  a leaky ReLU activation function at the hidden layer for the training.  %For training, sentences are processed one at a time; thus loss are calculated per sentence followed by an update of all network parameters.

\paragraph{Inference Policy:}
%We construct the graph $G(V,E)$ for a sentence $s$ using the approach discussed in previous sections. We also define the set $\mathcal{F}$ where $ \mathbf{f_{ij}} \in \mathcal{F}$ represents the PCRW feature vector for the directed edge $e_{ij} \in E$. %The edge $e_{ij}$ is the directed edge from the vertex $v_i$ to $v_j$, where $v_i,v_j \in V$. 

%For the sentence, the ground truth segment $S_{GT}$ will be an induced subgraph $G_{GT}(V_{GT},E_{GT})$ of $G$ with $V_{GT} \subseteq V$ and $E_{GT} \subseteq E $.  The gold standard will be a maximal clique in the graph as each node will be connected to every other node in $G_{GT}$. We, therefore, reduce our task of finding the segmentation solution to searching for the minimum weighted maximal clique from the graph $G$.

%Maximal clique finding is an NP-Hard problem and enumeration of all the maximal cliques require an exponential time algorithm \cite{tomita2006worst,bron1973algorithm}. Instead, 
For the maximal clique selection, we use a greedy heuristic approach inspired from Prim's algorithm \cite{prim1957shortest}. The policy is described in Algorithm \ref{algo1}. %for finding maximal cliques as our inference procedure. %which is originally used for finding Minimum Spanning Tree \cite{prim1957shortest}. %Our approach does not guarantee the enumeration of all the maximal cliques.  %Numerous approximation schemes has been proposed for finding sub-optimal solutions, of which \newcite{takahashi1980approximate} proposed a method by employing the Prim's algorithm used for finding the Minimum Spanning Tree \cite{prim1957shortest} in a graph. The work, primarily proposed for finding spanning tree in undirected graphs \cite{PLESNIK1992451} was extended to obtain arborescence for directed graphs \cite{voss1993worst}.

%In our work, like \newcite{takahashi1980approximate}, we use the Prim's algorithm as our search procedure for finding the relevant segments. We learn an energy based model which should ideally output an arborescence resembling that of the $G_{GT}$. %We are not concerned about the specific arborescence, but only about the set of nodes that the arborescence covers. 

\begin{algorithm}

  \SetAlgoLined
\label{algo1}
    \SetKwInOut{Input}{Input}
    \SetKwInOut{Output}{Output}

\For{each node $v_i$ in $V$ }{
   Initialize a graph $K_{i}(V_{K_{i}},E_{K_{i}})$ with $K_{i} = G$ such that $V_{K_{i}} = V$ and $E_{K_{i}}=E$. Initialise a vertex set $V_{T_{i}}$ with $v_{i}$ as the only element in it. Remove all the vertices which are conflicting with $v_{i}$ from $K_{i}$.

 Add the vertex $v_j \in (V_{K_{i}}-V_{T_{i}})$ to $V_{T_{i}}$, such that in $K_{i}$, the sum of edge weights for the edges starting from $v_j$ to all other vertices in $V_{T_{i}}$ is minimum.
    %\item From $E$, add all the edges with $v_j$ as source and for all $v_k$ in $V_{T_{i}}$ as target to $E_{T_{i}}$ . Since none of the vertexes currently in $V_{Temp}$ are conflicting with those in $V_{T_{i}}$, this construction will always result in a clique.

Remove all the vertexes which are conflicting with $v_j$ from $V_{K_{i}}$.

Repeat steps 3 - 4 till $V_{K_{i}} - V_{T_{i}} = \varnothing $    \caption{Greedy maximal clique selection heuristic}}

\end{algorithm}

\if{} 

For every node $v_{i} \in V$,
%\todo{I think the algo description below should change in accordance with your new figure.}
\begin{enumerate}[leftmargin=*]
%\vspace{-0.8em}
    \item Initialise a graph $K_{i}(V_{K_{i}},E_{K_{i}})$ with $K_{i} = G$ such that $V_{K_{i}} = V$ and $E_{K_{i}}=E$. Initialise a vertex set $V_{T_{i}}$ with $v_{i}$ as the only element in it. Remove all the vertices which are conflicting with $v_{i}$ from $K_{i}$.
    \vspace{-0.8em}
    \item  Add the vertex $v_j \in (V_{K_{i}}-V_{T_{i}})$ to $V_{T_{i}}$, such that in $K_{i}$, the sum of edge weights for the edges starting from $v_j$ to all other vertices in $V_{T_{i}}$ is minimum.
        \vspace{-0.8em}
    %\item From $E$, add all the edges with $v_j$ as source and for all $v_k$ in $V_{T_{i}}$ as target to $E_{T_{i}}$ . Since none of the vertexes currently in $V_{Temp}$ are conflicting with those in $V_{T_{i}}$, this construction will always result in a clique.
    \item Remove all the vertexes which are conflicting with $v_j$ from $V_{K_{i}}$.
    \vspace{-0.5em}
    \item Repeat steps 2 - 3 till $V_{K_{i}} - V_{T_{i}} = \varnothing$ 
    \vspace{-0.5em}
    
\end{enumerate}
\fi
%Figure \ref{ebg} shows our approach for obtaining one maximal clique. Here we show the clique selection approach for the input sequence in Figure \ref{chunk}. 

In Algorithm \ref{algo1}, we start the clique selection with a single node. %works as follows. We start with an arbitrary node that forms the clique to be expanded. 
At any given instance, we loop through the nodes in the graph which are not yet part of the clique. We add a vertex $v$ to the clique if the cumulative score of all the edges from $v$ to every vertex that is already in the clique is the minimum. We discard all the nodes which are conflicting with vertex $v$. ``Conflicting" nodes are any pair of nodes which are not connected by an edge between them. This follows from the construction of the graph $G$, as the non-connectivity between the nodes implies that they are proposed as alternative word suggestions in $G$. As guaranteed by our sentence graph construction, we obtain the maximal clique (\textit{exhaustive segmentation}) when there exist no more vertices to loop through. %The sentence graph construction guarantees that a maximal clique, i.e., an `\textit{exhaustive segmentation}' (\autoref{intro}) is always obtained using Algorithm \ref{algo1}. 
We perform this for every node in the graph $G$. From all the cliques so obtained we select the maximal clique with the least score. The approach does not guarantee enumeration of all the cliques, but it is guaranteed that every node will be covered by at least one maximal clique. 
The heuristic can be seen as a means of sampling some potential minimum energy maximal cliques for the learning task. %of the energy based model
Energy based models do not require proper normalisation of the solution space \cite{lecun2006tutorial}, a choice that enables the use of the heuristic. 

%\newcite{mcdonald2005simple} previously used Bron-Kerbosch algorithm \cite{tomita2006worst,bron1973algorithm}, an exponential time ($O(3^{\frac{n}{3}})$) clique enumeration algorithm, for a relation extraction task. To make their graphs  sparsely connected, the authors pruned the edges in their originally fully connected graph before the algorithm was used. 
During inference, the greedy clique selection heuristic is performed for every node in $G$. Though the run-time for this inference is polynomial, it can still be computationally expensive.  %in terms of the number of nodes in the graph.\iffalse, it still can be at best $O(n^3)$ which is computationally quite expensive.\fi On the other hand, bron-keroshc algorithm used to enumerate all the cliques in a graph \cite{tomita2006worst,bron1973algorithm,mcdonald2005simple} runs in an exponential time of . 
But, in practice we find that our inference procedure results in faster output for graphs with $>19$ nodes in comparison to the exponential time Bron-Kerbosch algorithm~\cite{tomita2006worst,bron1973algorithm} for clique enumeration \cite{mcdonald2005simple}. We further improve the run time of our inference procedure by paralleling the clique selection procedure for each node on a separate thread. %\iffalse. Since the infernece look for cluique starting from every node, this step can be parallelised\fi as against the exhaustive enumeration algorithm.  %The clique  so obtained denotes a phonetically valid solution. 

\vspace{-0.5em}

\if{}
We represent our corpus also as a graph, $G_{glob}(V_{glob},E_{glob})$. For the sentences in the corpus, we not only use the segmented word forms but also the corresponding morphological analyses of the words. Hence the nodes in $G_{glob}$ can be of different types. %Due to the free word nature of Sanskrit, we keep co-occurrence of two words (or other types) in the sentence as the sole criteria for edge formation in the network.
If two nodes co-occur in the corpus graph, we form 2 directed edges and assign the co-occurrence probability as the edge weights. Now, relying solely on co-occurrence may lead to noisy contexts. So we only consider the connectivity between the nodes in the corpus graph via a set of filtered paths. We use the path ranking algorithm in PCRW \cite{lao2010relational}, to identify the typed paths in $G_{Glob}$, which are useful for the task. For every pair of candidate nodes in the sentence graph $G$, each typed path assigns a non-negative real valued score based on the information from the corpus graph. The set of scores hence forms the feature vector for the edge in the sentence graph.

%To be specific, the features correspond to metapaths obtained from a large corpus. We construct a graph $G_{Glob}$ From a large corpus where the nodes are of heterogeneous type. 
 %For every edge in the sentence graph $G$, we perform random walk over the corpus graph $G_{Glob}$, via each of the filtered typed paths. For an edge, the scores so obtained via each of the metapaths are then represented as a vector where each dimension of the vector corresponds to a unique typed path. 
 The energy based model is then fed with the graph structure $G$ and the set of edge vectors. The model searches for the minimum energy maximal clique of the sentence graph $G$. Figure \ref{archi} shows the architecture of our system.    %We propose a framework for automatic generation of embeddings with interpretable dimensions using PCRW.

%For a given sentence, we obtain all of its phonetically valid segmentations along with the lemma and morphological information for the segmented candidate words from the ``Sanskrit Heritage Reader''.  %Given a sequence of characters, which we refer to as a `chunk', it implies that there exist at least one word in the chunk. Figure \ref{} shows a sample output from the SKT, where it enumerates all possible outputs for a given chunk of characters. As shown in the Figure \ref{}, SKT also provides the lemma and morphological information (henceforth to be referred to as morph) for a given word. 

\section{Candidate Space Generation and Sentence Graph Construction}

%The aim of the word segmentation task is to obtain a valid word level segmentation for a given input sentence. 
%We obtain all the phonetically valid segmentations for a sentence using the Sanskrit Heritage Reader (SHR)\footnote{http://sanskrit.inria.fr/DICO/reader.fr.html}, a shallow parser for Sanskrit. 

Figure \ref{herit} shows the possible segments and the desired segmentation solution for a sentence, `{\sl satya{\d m}br\=uy\=atpriya{\d m}br\=uy\=annabr\=uy\=atsatyamapriya\-{\d m}priya{\d m}can\=an{\d r}tambr\=u\-y\=ade{\d s}adharma{\d h}san\=atana{\d h}}'\footnote{This is a well-known saying ({\sl subh\=a{\d s}itam}) in Sanskrit: ``One should tell the truth, one should say kind words; one should neither tell harsh truths, nor flattering lies; this is a rule for all times.''}. The Sanskrit Heritage Reader essentially shows all the unique segments that are part of at least one segmentation solution.  For example, in Figure \ref{herit}, the word `\textit{sati}', numbered as 9, is part of 264 of possible 1056 segmentation solutions. %The sentence in total has 1056 possible segmentations as solutions. 
We call such a representation of the segments as a shared forest representation.

%We rely on the Sanskrit heritage reader's lexical juncture system \cite{goyal2016design} to decide whether two nodes are conflicting or not. 

The Sanskrit Heritage Reader uses finite state methods in the form of a lexical juncture system to obtain the possible segments for a given sentence.  We follow the definitions as used in \newcite{goyal2016design} and it is recommended to refer the work for a detailed review of the system.
 
A {\sl lexical juncture system} on a finite alphabet $\Sigma$ is composed of a finite set of words $L\subseteq \Sigma^*$ 
and a finite set $R$ of rewrite rules of the form $u|v\rightarrow f/x\_\_$ \cite{kk}, with $x,v,f\in\Sigma^*$ and
$u\in\Sigma^+$. In this formalization, $\Sigma$ is the set of phonemes, $R$ is the set of sandhi rules, and $L$ is the vocabulary as a set of lexical items. Though every entry $z \in L$  is an inflected word form, it additionally contains the morphological analysis of the word as well. For clarity, we will denote every entry $z$ additionally as a 3-tuple $(l,m,w)$, where $l$ denotes the root of the word, $m$ denotes the morphological class of the word, $w$ denotes the inflected word form generated from $l$ and $m$. %and $k$ denotes the position at which the word $f$ begins in the sentence s. %Now $l_{i}$ represents the inflected word form generated from the 3 tuple $(l,m,k)$.
Given a sentence $s$, a valid segmentation solution/sandhi analysis, $S_{i}$, can be seen as a sequence 
$ \lbr z_1,\sigma_1,k_1 \rbr ;... \lbr z_p,\sigma_p , k_p \rbr $. Here, $\lbr z_j,\sigma_j,k_j\rbr $ is a segment with $z_j\in L$, $k_j \in \mathbb{N}$ denotes the position at which the word $w_j$ begins in the sentence $s$ and $\sigma_{j} = \lsq x_{j} \rsq u_j|v_j \rightarrow f_j \in R$ 
for $(1\leq j\leq p)$.\iffalse , $v_p=\epsilon$ and $v_j=\epsilon$ for $j<p$ only if
$\sigma_j=o$, subject to the matching conditions:
$z_j=v_{j-1} y_j x_j u_j$ for some $y_j\in\Sigma^*$ 
for all $(1\leq j\leq p)$, where by convention $v_0=\epsilon$.  
Finally $s=s_1 ... s_p$ with
$s_j=y_j x_j f_j$ for $(1\leq j\leq p)$, $\epsilon$ denotes the empty word.\fi
%We also say that such a sequence is an {\sl analysis} of the solution
%word $s$.

For $s$, there can be multiple possible \textit{sandhi} analyses. Let $\mathcal{S}$ be the set of all such possible analyses for $s$. We find a shared forest representation of all such \textit{sandhi} analyses
\vspace{-0.8em}
$$D(\mathcal{S}) = \bigcup_{S_{i}\in\mathcal{S}} {S_{i}}\vspace{-0.7em}$$ 

A segment $\lbr z_j,\sigma_j,k_j\rbr \in D(\mathcal{S}) $, iff $ \lbr z_j,\sigma_j,k_j\rbr $ exists in at least one $S_{i}$. Two segments $\lbr z_i,\sigma_i,k_i\rbr$ and   $\lbr z_j,\sigma_j,k_j\rbr$ are said to be `conflicting' if $ k_{i} \leq k_{j} < k_{i} + |w_{i}| - 1$ or $ k_{j} \leq k_{i} < k_{j} + |w_{j}| - 1$. No two conflicting segments exist in a valid segmentation solution.% including the ground truth sandhi analysis $S_{GT}$.

\subsection*{Converting the candidate space into graph}
We convert the shared forest representation of the candidate segments into a sentence graph $G(V,E)$. For the graph $G$, a node $v \in V$ is a segment $ \lbr z_j,\sigma_j,k_j\rbr $, where $z_j$ is a 3-tuple $(l,m,w)$. This representation scheme for the segment is indispensable for the task. For example, in Figure \ref{herit}a, the nodes 1, 2 and 3 differ from each other only based on the morphological tag they carry. It is represented using the $m$ attribute of the 3-tuple $(l,m,w)$. Similarly nodes 1 and 11 differ from each other only based on their position information represented by $k_j$ in the segment.  %shows three different nodes, marked as 1, 2 and 3, corresponding to each morphological class for {\sl satyam}. Second, the same word might be repeated twice in the sentence, but in a different position. In Sanskrit, words especially verbs are often repeated to lay emphasis on the action performed \cite{krishna2016segmentation}. We need to keep the position information explicitly to distinguish the candidate segments from one another in such cases. For example, in Figure \ref{herit}a, nodes 11, 12 and 13 differ from nodes 1, 2 and 3 respectively only by virtue of their positions in the sentence. 
%Two nodes are `\textit{conflicting}' if they have an overlap in the position relative to the sentence and the overlapped portion does not adhere to any of the rules that follow \textit{sandhi}. 
 %due to two major reasons. First, the morphological analysis of a given inflected word-form in Sanskrit may be ambiguous and can lead to multiple morphological classes. In Figure \ref{herit}a, the word `{\sl satyam}' is an inflection of the lemma `{\sl satya}', and the inflection is used to denote three different morphological classes. Humans rely on syntactic or semantic context to understand the intended morphological class in such cases. We treat each inflection as a separate candidate. Figure \ref{herit}a shows three different nodes, marked as 1, 2 and 3, corresponding to each morphological class for {\sl satyam}. Second, the same word might be repeated twice in the sentence, but in a different position. In Sanskrit, words especially verbs are often repeated to lay emphasis on the action performed \cite{krishna2016segmentation}. We need to keep the position information explicitly to distinguish the candidate segments from one another in such cases. For example, in Figure \ref{herit}a, nodes 11, 12 and 13 differ from nodes 1, 2 and 3 respectively only by virtue of their positions in the sentence. 
%Two nodes are `\textit{conflicting}' if they have an overlap in the position relative to the sentence and the overlapped portion does not adhere to any of the rules that follow \textit{sandhi}. 

\begin{figure}[!htb]
\hspace*{1in}

\centering
\vspace{-0.2em}
\minipage{\columnwidth}
\hspace*{1cm}\includegraphics[width=0.6\columnwidth]{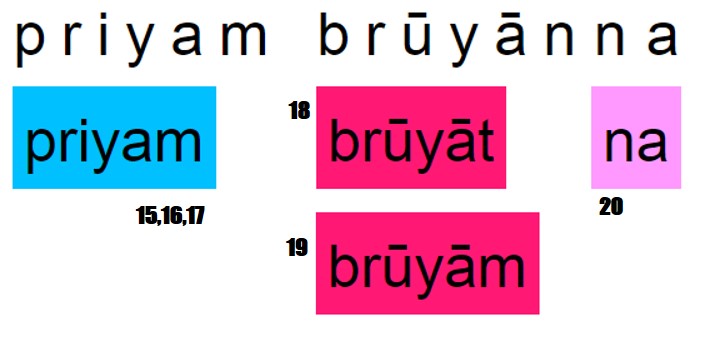}
\vspace{-6mm}
\caption{A subsequence of sentence from Figure \ref{herit}}
\label{chunk}
\endminipage\hfill
\vspace{-4mm}
\end{figure}

\begin{figure}[!htb]
%\hspace*{0.2in}
\centering
\vspace{-0.2em}
\minipage{\columnwidth}
\hspace*{1.5cm}\includegraphics[width=0.5\columnwidth]{images/InitG.jpg}
\vspace{-3mm}
\caption{Sentence Graph constructed for the sub-sequence in Fig. \ref{chunk}. Edges are bidirectional. For better illustration of construction details, node 17 is omitted.} %Node 17 will have edges to all other nodes other than nodes 15 and 17}
\label{initG}
\endminipage\hfill
\vspace{-3mm}
\end{figure}

\fi

\section{Feature Generation for the Edges}
\label{PCRW}

Given two non-conflicting nodes in $G$, there exists a pair of directed edges, one each in either direction.
For every edge in the sentence graph $G$, we need to generate features that capture the distributional information between the candidate nodes that the edge connects. Similar in spirit to \newcite{bilmes2003factored} and \newcite{krishna2016segmentation}, we condition the distributional information based on different morphological constraints to enrich the context.  %We use a morphologically tagged corpus in Sanskrit for this. 
%We can obtain a language model that captures the sentence level co-occurrence information between words From a corpus, given the language follows a free word order. %But, considering co-occurrence between the words as the sole criteria can be noisy. %For example, the adjective of the subject in  a  sentence  may  not  be  of  any  relevance  to the object in the sentence. 
The distributional information is obtained from a morphologically tagged corpus $\mathcal{C}$. Let $\mathcal{V}_{w}$, $\mathcal{V}_{m}$ and $\mathcal{V}_{r}$ be the vocabulary of the inflected word forms, morphological classes and the roots respectively in the corpus. Let $\mathcal{V} = \mathcal{V}_{w} \cup \mathcal{V}_{m} \cup \mathcal{V}_{r}$. For each $n_i, n_j \in \mathcal{V}$, the conditional probability is calculated as  $P_{co}(n_j|n_i) = \frac{count(n_j,n_i)}{count(n_i)}$. Here $count(\cdot)$ represents the count of co-occurrence between the entries in the corpus. Also, let $MC$ be the set of morphological constraints used for conditioning the distributional information. Now, for each $n_i,n_j \in \mathcal{V}$ and each $g_k \in MC$, we can obtain the feature value as follows:
\begin{equation*}
\label{coProb}
P_{g_k}(n_{j}|n_i) = -log ( P_{co}(n_j|g_k)\times P_{co}(g_k|n_i))
\end{equation*}

We use the following scheme for feature generation. A node in $G$ essentially contains three attributes, namely, the root, the morphological class and the inflected word form. A feature uses corpus evidence of exactly one of the three attributes.  For instance, consider two candidate nodes $o_1, o_2$ in $G$ with $(o_{1w},o_{1m},o_{1r})$ and $(o_{2w},o_{2m},o_{2r})$  as the respective 3-tuple attributes. Now, one such possible feature value for the edge from $o_1$ to $o_2$ can be calculated as $P_{g_1}(o_{1r}|o_{2w})$ where $g_1 \in MC$ and $o_{1r},o_{2w} \in \mathcal{V}$. Hence, features for a directed edge connecting two different nodes in $G$ can be generated in $3 \times |MC| \times 3$ ways.

We automate the process of feature generation and feature selection using the framework of Path Ranking Algorithm \cite{lao2010relational}. Formalising our approach using PRA leads to an efficient and scalable implementation of our scheme. In PRA, enumerating and generating all the possible features needs to be performed only for a sampled set of data pairs from the corpus. By using a supervised feature selection approach, a relevant sub-set of features is filtered. This is a one-time process \cite{gardner-mitchell:2015:EMNLP}. During inference, the feature values are computed only for the filtered features. \footnote{The edge vector formation is explained in terms of Metapaths \cite{sun:2010:POSTERS} in \S 3 of the Supplementary.}

 %Path Constrained Random Walks (PCRW) \cite{lao2010relational} ; But similar to \newcite{bilmes2003factored} the features were  handcrafted. 
%Here, we define $GC$ as the set of all combinations of grammatical categories in Sanskrit. 

%Thus, given a pair of entries from the $\mathcal{V}$, we can generate $|GC|$ different feature values. %But, a node in the sentence graph $G$ represents 3 entries, the lemma, morphological class and the inflected word form. Hence given an edge in $G$ there are $9 \times |GC|$ features possible. 

\textbf{Morphological Constraints:} $MC$ is defined as the set of grammatical category combinations, each combination falling into one of the following two descriptions. %where only those combiantios between the categories that are combatible between each other are combined.  following two types. 
a) {\sl Complete combination}, i.e., a morphological class -- In Sanskrit, similar to Czech \cite{smith-smith-tromble:2005:HLTEMNLP}, a morphological class represents a certain combination of grammatical categories. For instance, a noun is represented by case, gender and number. Hence, the combination `genitive-masculine-singular' forms a morphological class. b) {\sl Partial combination} - A combination of grammatical categories, which can form a morphological class by adding one or more categories to it. For instance, `genitive-masculine' is a partial combination that denotes all the possible (three) morphological classes which differ from each other only in terms of the category `number'. However, `genitive-present tense' is not a `valid' combination as it can never form a valid morphological class. The evidence for a partial combination in the corpus $\mathcal{C}$ can be obtained by summing the evidence of all the morphological classes which it denotes. We obtain a total of 528 morphological constraints. % which differ to each other only on the basis of the missing cateogry(ies).
%This information encoded in $MC$ is readily available on any introductory work on Sanskrit Grammar, hence the effort involved is trivial.
A filtered set of 1500 features (out of 4752) is used in our model. %With 528 constraints, 4752 features are possible. 
Mutual Information Regression  \cite{kraskov2004estimating} with the word to word co-occurrence probability as label is used for feature selection.\footnote{For different settings we experimented with, for the vector generation, refer to \S4 of the supplementary material.} %Currently every feature is conditioned exactly on one morphological constraint. We find that increasing the count of constraints per feature leads to decrease in performance.

\if{} 
\begin{table}[!htb]
\vspace{-2mm}
\centering
\begin{small}
\begin{tabular}{|l|l|lll}
\cline{1-2}
\textbf{Type}        & \textbf{No. of nodes} \\ \cline{1-2}
Grammatical Category & 310              &  &  &  \\ \cline{1-2}
root                & 66,914           &  &  &  \\ \cline{1-2}
Word                 & 217,535         &  &  &  \\ \cline{1-2}
Morphological Class  & 218              &  &  &  \\ \cline{1-2}
\end{tabular}
\end{small}
\caption{Node types and the number of corresponding nodes in $G_{Glob}$}
\label{nodeOcc}
\vspace{-5mm}
\end{table}

\fi
\if{}
The score for a typed path can be calculated as,
\vspace{-1em}
\begin{equation}
\label{coProb3}
-log_{10} ( \prod_{i=1}^{n-1} P_{co}(v_{i+1}|v_{i}) )   
\vspace{-0.5em}
\end{equation}

where nodes $v_{1}\ldots, v_{n} \in V_{Glob}$ and $v_{i}$ follows the type for the corresponding position in the typed path.
\fi

\section{Experiments}
\label{exp}

\paragraph{Dataset:} 
We use the Digital Corpus of Sanskrit (DCS) \cite{dcsOliver}, a morphologically tagged corpus of Sanskrit, for all our experiments. DCS contains digitised works from periods that span over 3000 years and contain constructions written in prose or poetry. %\iffalse Our primary motivation to design a model that is agnostic to ordering is the challenge in identifying the writing style followed in a construction, as there are many works where both prose and poetry are used together and is a laborious task to identify the writing style by itself.\fi 
Identifying sentence boundaries in Sanskrit constructions is a challenge of its own \cite{hellwig:2016:COLING}. DCS currently has split the corpus into more than 560,000 text lines, all of which need not be following explicit sentence boundaries.  %The corpus contains manuscripts which are stylistically, topically and chronologically diverse. %and contains more than 66,000 unique lemmas and 3,200,000 tokens.
\newcite{krishna2016segmentation} identify 350,000 constructions from the DCS fit for the segmentation task. They use a separate set of 9,577 constructions from the DCS, called as `DCS10k', and use it as the test set.  They ignore the remaining text lines from DCS due to ambiguities either in the provided tagging or alignment with SHR \cite{krishna-satuluri-goyal:2017:LaTeCH-CLfL}. %\iffalse of which about 40,000 of the textlines were ignored due to ambiguity in the tagging provided in DCS. The others were ignored due to challenges in aligning them with the nomenclature and conventions used in Sanskrit Heritage reader\fi.  
We use the 350,000 constructions used in \newcite{krishna2016segmentation} as the corpus $\mathcal{C}$ (\S \ref{PCRW}) for the generation of our edge vectors. `DCS10k' was  neither used in the training of our model, nor in the generation of edge vectors. \newcite{reddy2018seg} report their results on a subset of DCS10k containing 4,200 sentences, which we will refer to as `DCS4k'.

We experiment with the following systems:

\noindent \textbf{SupervisedPCRW:} Proposed in \newcite{krishna2016segmentation}, this model also uses the graph output from SHR. Using PCRW \cite{lao2010relational,Meng:2015:DML:2736277.2741123}, feature vectors for edges are generated using hand-crafted morphological constraints. % as detailed in \newcite{krishna2016segmentation}. %\iffalse The weight vector for the feature vectors are learnt using a logistic regression classifier as performed in \newcite{lao2010relational}.\fi 
Starting with the longest word in the graph, the prediction is performed by greedily selecting the candidates as per the edge weights. %The final edge weight is determined by the dot product between the weight vector and the feature vector. But, in 3 of the components the edge vectors are modified as per the sentnece graph structure G, while the other components are pre-computed based on a corpus co-occurrence. 
\\
\noindent \textbf{EdgeGraphCRF:} This is a second order CRF Model \cite{muller2014pystruct,ishikawa2011transformation} which uses the sentence graph structure $G$ as the input to the system. Every node is represented with fastText  \cite{bojanowski2016enriching} word embeddings trained under default settings. The edges are featurised with the PRA vectors (\S \ref{PCRW}).  %Since, this is a second order model the dependendcies between other edges are not taken into consideration, while making the prediction. %, as it was performing better than max-product belief propagation approach. 
We used 1-slack structured SVM for training. %The values for regularisation constant and convergence tolerance were set to 0.05.
For the binary class problem, QPBO \cite{rother2007optimizing} inference approach provided the best results.

\noindent \textbf{Seq2Seq} - \newcite{reddy2018seg} uses an Encoder-Decoder framework with LSTM cells for the segmentation task. We consider two models from the work, namely, `\textit{segSeq2Seq}' and `\textit{attnSegSeq2seq}' as our baselines. The later which uses attention \cite{bahdanau2014neural} is the current state of the art in Sanskrit word segmentation.
   
\noindent \textbf{Lattice-EBM:}  An energy based sequence labelling model, where the input is a lattice~\cite{wolf1977hwim} similar to that of \newcite{kudo2006mecab}. The model can be seen as a special case of Graph Transformer Networks \cite{lecun1998gradient,lecun2007energy}. In the lattice structure, the candidate links only to its adjacent nodes in an exhaustive segmentation. We also generate edge vectors for the dummy nodes that act as the start and end markers in the lattice. During prediction, we have to find the best path from the lattice which minimises the sentence score. Here, we consider two variants of Lattice-EBM. \textit{L-EBM-Vanilla} uses the discriminative forward training approach \cite{Collobert:2011:NLP:1953048.2078186} with the standard hinge loss. The second variant \textit{L-EBM-Beam}, uses multi-margin loss \cite{edunov2017classical}, instead of the hinge loss. Here, we employ beam search to generate multiple candidates as required by the loss. 
 
\noindent \textbf{Tree-EBM:} The baseline model works exactly the same as the proposed model where the only difference between both is in the inference algorithm used. Tree-EBM has an inference that searches for a Steiner Tree  \cite{takahashi1980approximate} from the input graph $G(V,E)$, the structure of which is described in \S \ref{inGraph}.\footnote{The inference procedure is given in \S2 of the supplementary material} The inference procedure outputs a spanning tree that covers a subset of the nodes from $G$. This raises a challenge while estimating the loss as, unlike in the case of a clique, there can be multiple rooted tress that spans the subset of nodes in the ground truth. In this model, we augment the loss $L_{tree}$ such that the Steiner tree with the least energy that spans the nodes in ground truth is chosen.
 
 \begin{dmath*}
\begin{small}
L_{tree}=max(0, \argmin_{T_m \in A^{G} }\mathcal{S}(T_{m}) - \argmin_{T_i\in A^{Q}}(\mathcal{S}(T_{i}) - \Delta(T_{i}))
\end{small}
\end{dmath*}
 
 Here $A^{G}$ represents set of all the trees that spans the nodes in the ground truth.
 
\noindent \textbf{Clique-EBM:} The proposed model. The EBM model uses the maximal clique selection heuristic for the inference.

%In the dataset, every sentence is labelled with the segmentations, lemma and morphological information about each of the individual segmented word. We filter out 350,000 of the sentences to form the network $G_{glob}$, which is used for the construction of PCRW vectors. %The held-out data used in \newcite{krishna2016segmentation}, a set of 9,576 sentences, henceforth referred to as `\textit{DCS10K}', is used for testing and is neither part of our training nor was used in construction of $G_{glob}$. We discarded about 40,000 sentences from DCS due to incomplete information in terms of analysis for those sentences either in terms of lemma or morphological information. 

\noindent \textbf{Tasks and Evaluation Measures:}  We use the macro-averaged Precision (P), Recall (R), F1-score (F)  and also the percentage of sentences with perfect matching (PM) as our evaluation metrics. We evaluate the competing systems on the following two different tasks. %as described below. We use  for both the tasks. We also use  as an evaluation criteria.

{\sl Word Prediction Task \textbf{(WPT)}} - The word segmentation task is evaluated based on the correctness of the inflected word forms predicted.  This was used in \newcite{krishna2016segmentation}. %It is possible that the morphological class need not be correctly predicted.%Since, every node $v$ is a 3-tuple, it is possible that the predicted node, the final word form and lemma information is correct but the morphological class is incorrect. So, we can claim that the predicted word form is correct, but no additional information from the system can be reliably obtained, other than the correctness of lemma. In this setting, the checking for correctness of lemma alone suffices. 

{\sl Word++ Prediction Task \textbf{(WP3T)}} - This is a stricter metric for the evaluation of the joint task of segmentation and morphological tag prediction. It evaluates the correctness of each of the inflected word form, lemma and its morphological tag. %morpheme (root, morphological tag and the inflection).%, and we can obtain reliable information about the morphological properties of a word, which will be beneficial for further syntactic and semantic level tasks. 

%We report all our scores based on the macro-averaged precision and recall over the test-set for the word prediction task unless otherwise stated.

\renewcommand{\thetable}{1.\Alph{table}}
\begin{table}[!htb]
%   \resizebox{10cm}{!}{
    \begin{minipage}{.5\textwidth}
        \centering
    \begin{small}
    \begin{tabular}{|c|c|c|c|c|c|}
\hline
No: &System              & P                & R                & F                & PM              \\ \hline

1&segSeq2Seq          & 73.44            & 73.04            & 73.24            &   29.2              \\ \hline
2&SupervisedPCRW      & 76.30            & 79.47            & 77.85            & 38.64           \\ \hline
3&EdgeGraphCRF        & 79.27             & 81.6            & 80.42            & 35.91           \\ \hline
4&L-EBM-Vanilla      & 83.12           & 85.49            & 84.29            &        53.62      \\ \hline
5&L-EBM-Beam      & 86.38           & 85.77            & 86.07            &        60.32      \\ \hline

6&AttnsegSeq2Seq      & 90.77            & 90.3            & 90.53            &             55.99    \\ \hline

7&Tree-EBM      & 89.44           & 92.35            & 90.87            &        61.72      \\ \hline

\textbf{8}&\textbf{Clique-EBM} & \textbf{96.18} & \textbf{97.67} & \textbf{96.92} & \textbf{78.83} \\ \hline
\end{tabular}
\end{small}
% 90.88	82.35	87.96	79.74	89.4
\vspace{-3mm}
\caption{WPT on DCS4k}
\label{tab:accuracy} 
    \end{minipage}
    \begin{minipage}{.5\textwidth}
        \centering
        \vspace{3mm}
\begin{small}
\begin{tabular}{|l|c|c|c|c|}
\hline
System       & P     & R     & F     & PM    \\ \hline
EdgeGraphCRF &  76.69     &  78.74      & 77.7  & 31.82 \\ \hline

LatticeEBM-Vanilla &  76.88     &  74.76      & 75.8  & 27.49 \\ \hline

LatticeEBM-Beam &  79.41     &  77.98      & 78.69  & 31.57 \\ \hline

Tree-EBM &  82.35     &  79.74      & 81.02  & 32.88 \\ \hline
\textbf{Clique-EBM } & \textbf{91.35} & \textbf{89.57} & \textbf{90.45} &  \textbf{55.78}       \\ \hline

\end{tabular}
\end{small}
\caption{WP3T on DCS10k}
\label{wppResults}
    \end{minipage}
%}  
\setcounter{table}{0}
\renewcommand{\thetable}{\arabic{table}}
\caption{Performance evaluation of the competing systems in ascending order of their F-Score.}
\label{allResults}
\vspace{-6mm}
\end{table}

\subsection{Results}

%We compare our system PCRW-EBM with the best performing model of \newcite{krishna2016segmentation}, named as SupervisedPCRW and two other structured prediction systems GraphCRF and EdgeFeatureGraphCRF. 

Table \ref{allResults} provides the results for the best performing configurations for each of the systems. The results for WPT on DCS4k and WP3T on DCS10k for each of the systems are shown in Tables \ref{tab:accuracy} and \ref{wppResults}. The proposed model, Clique-EBM (System 8), outperforms all the other models across all the 4 metrics on both the tasks. Clique-EBM shows an improvement of 7.06\% and 40.79\% in terms of F-score and {\sl perfect matching} from the current state of the art (System 6) in WPT. Currently there exists no system that predicts the morphological tags for a given word in Sanskrit. %A phonetically valid split is guaranteed by virtue of the infernece procedure designed in case of both the EBM models, 
For WP3T, Clique-EBM has shown a percentage increase of  11.64\% and 69.65\%  in F-Score and the perfect matching score from Tree-EBM, the next best system.

All the systems except 1 and 6 in Table \ref{tab:accuracy} use the linguistically refined output from SHR as their search space to predict the final solution. Out of which 3, 4, 5, 7 and 8 use the edge vectors, which encodes the morphological constraints refined using PRA \cite{lao2010relational},  generated by the method discussed in \S \ref{PCRW}. As a result these systems require $\textless10\%$ training data than required by system 6. System 3 was trained on 10,000 sentences, while systems 4 and 5 were trained on 9,000 sentences after which the models got saturated. Systems 7 and 8  were trained on 8,200 sentences which is 7.66\% of the training data (107,000) used in System 6. In terms of training time, \newcite{reddy2018seg} reports a training time of 12 hours on a GPU machine, while systems 7 and 8 take a training time of 8 hours on an Intel Xeon CPU based machine with 24 cores %48 threads (2 CPU x 12 cores x 2 threads) 
and 256 GB RAM.\footnote{Please refer \S 4 of the supplementary for wall time analysis. System 6, when trained on this CPU based system, did not converge even after 15 hours of training.} %It takes less than 30 minutes to extract the SHR output for all the 8200 sentences used in training. 
For systems 4 and 5 it takes roughly 4 hours to train on the same machine. %All our wall time analysis is reported based on training on XX system. 
There was no training involved for the prediction of segmentations in system 2.
%None of the systems other than 8 outperforms system 6 in all the 4 metrics for WPT. Systems 5 and 7 outperforms 6 in at least one of the metrics. The differences arise from the different model architectures and the training used. 

In systems 4 and 5, %a standard lattice parsing model, the edges are formed only to position wise immediate nodes in an exhaustive segmentation. 
the inference is performed sequentially from left to right\footnote{We also tried reversing the input effectively enabling the right to left  direction but the results were worse than the reported system by an F-Score of 3.}. The use of beam search with multi margin in System 5 resulted in marginal improvements ($\textless 2$) in terms of F-Score to that of system 4. Further, the improvement in the results saturated after a beam size of 128. %all results within a two point difference in F-Score. %The best results are reported in Table
System 3 being a second order CRF model \cite{ishikawa2011transformation}, does not take the entire sentence context into account. In fact, about  85.16\% of the sentences predicted by the model from DCS10K do not correspond to an `{\sl exhaustive segmentation}'. %The predictions either contain conflicting candidates or some portion corresponding to the input is missing. 
Prediction of an `\textit{exhaustive segmentation}' is guaranteed in all our EBM models 4, 5, 7 and 8 (also in system 2) by virtue of the inference procedure we use. Both systems 7 and 8, along with System 6 which uses attention, consider the entire input context when making each prediction. System 8 considers all the pairwise potentials between the nodes while making a prediction, but System 7 does not (Steiner tree vs. maximal clique).

\begin{figure}[!htb]
\centering
\vspace{-2mm}
  \includegraphics[width=\columnwidth]{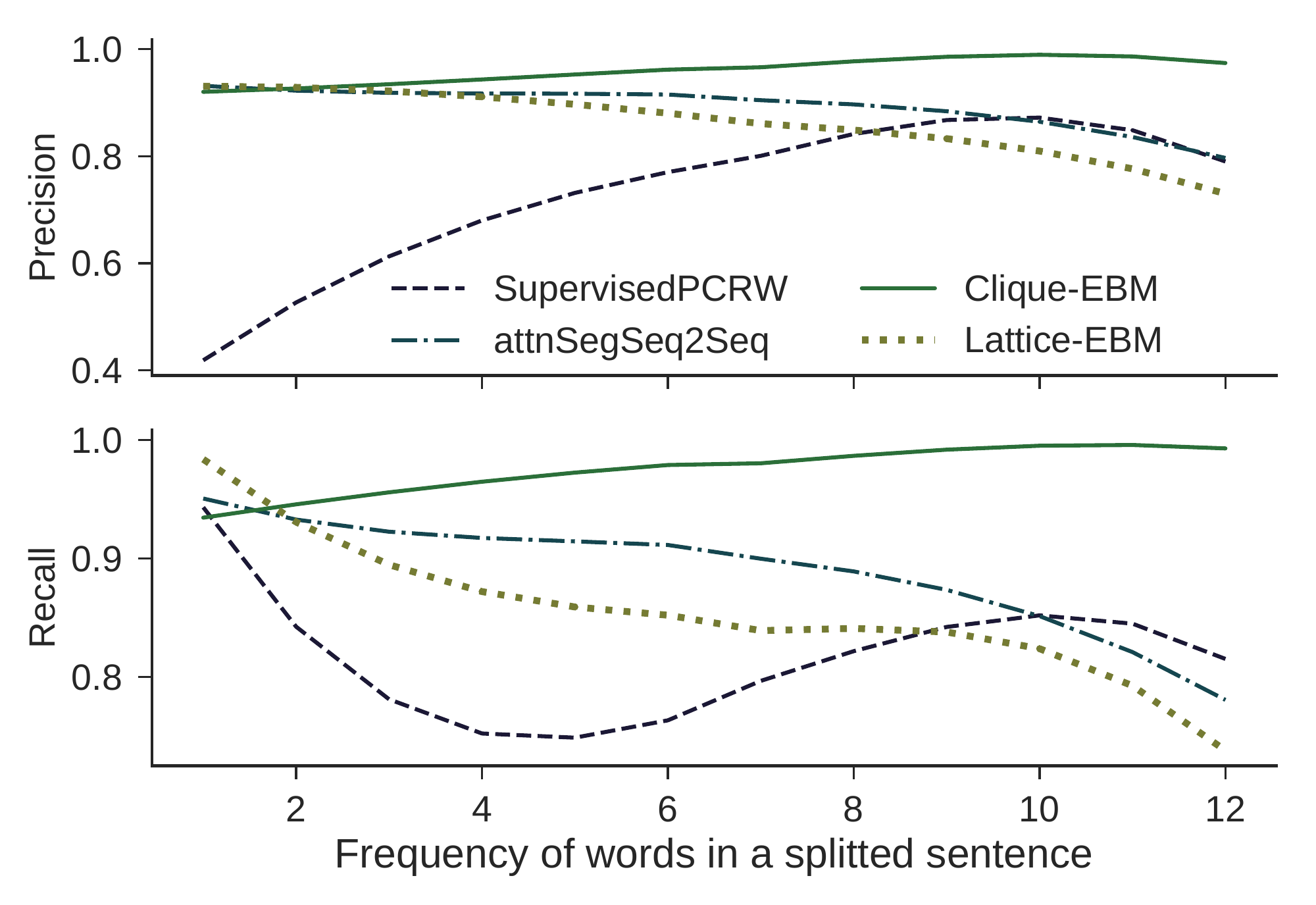}
  \vspace{-7mm}

  \caption{Performance of the competing systems for DCS4k grouped on the word counts in the ground-truth}
  \label{lenPR}
  \vspace{-5mm}
\end{figure}

Figure \ref{lenPR} reports the performance of the systems 2, 5, 6 and 8 where the sentences in DCS4k\footnote{sentences with length more than 12 words are not shown as such sentences appear less than 10 times in DCS4k.} are categorised based on the number of words presented in the segmented ground-truth solution. Our proposed system Clique-EBM performs the best across all the lengths with an exception towards shorter constructions of 2 words or less. Interestingly, both the sequential models (systems 5 and 6) show a decreasing trend as the number of words increases, while the Clique-EBM model shows an increasing trend with a larger length, which might indicate that more context helps the model. In fact, the greedy yet non-sequential approach used in \newcite{krishna2016segmentation} outperforms both the sequential models at longer lengths. The average length of a sentence in DCS is 6.7 \cite{krishna2016segmentation}, the share of sentences with seven or more words is 62.78\%. 

\if{}

\begin{figure}[!htb]
\vspace{-1mm}
\centering

\minipage{\columnwidth}
  \includegraphics[width=\textwidth]{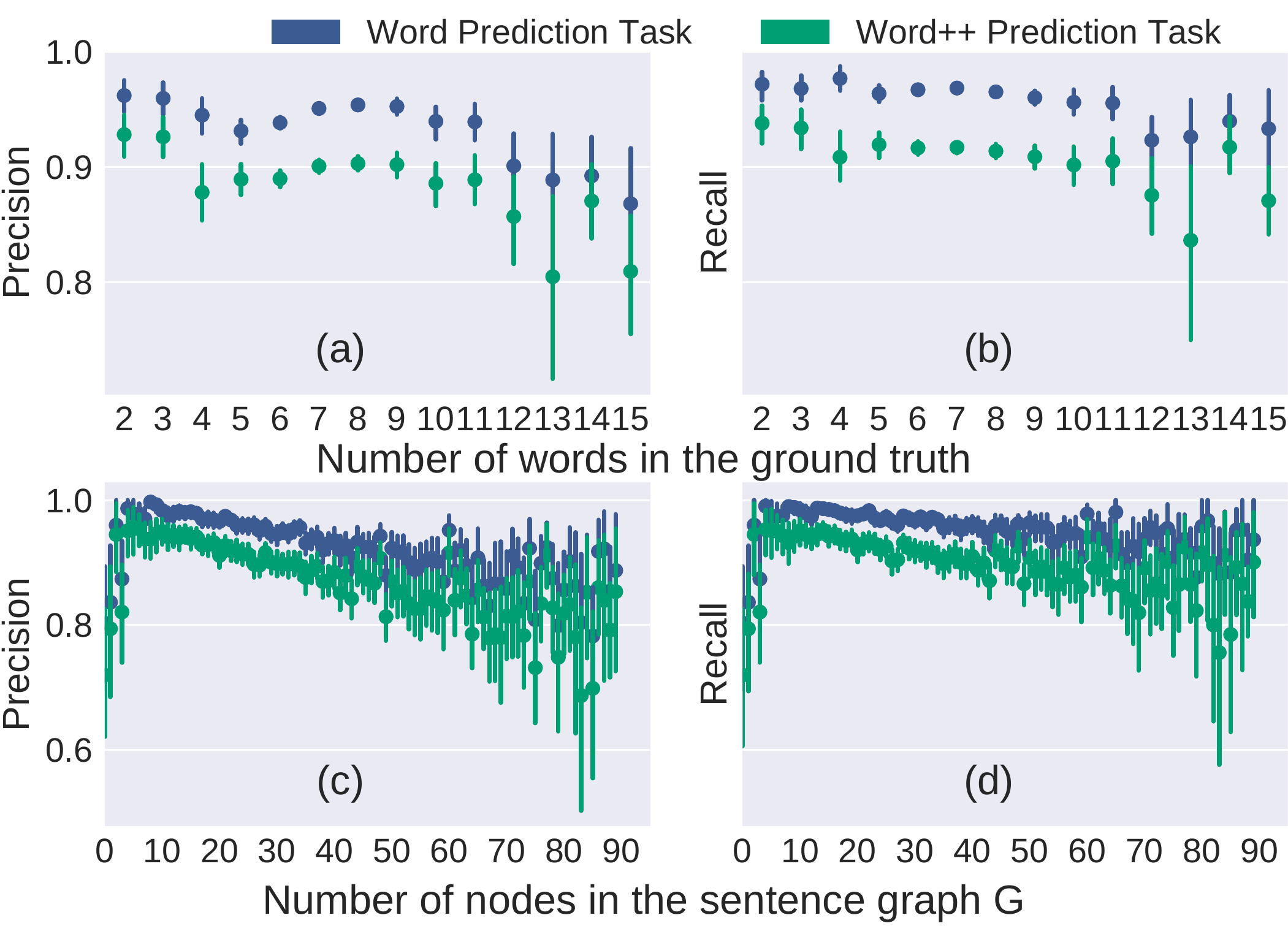}
  \vspace{-1.8em}
  \caption{Performance of the system in terms of Precision and Recall for entries in DCS10K grouped over  a-b) Number of words in ground truth c-d) Number of nodes in input graph G. }

  \label{PRF}
\endminipage\hfill
\vspace{-2mm}
\end{figure}

\begin{figure}[!htb]
\centering
\vspace{-2mm}
  \includegraphics[width=0.8\columnwidth]{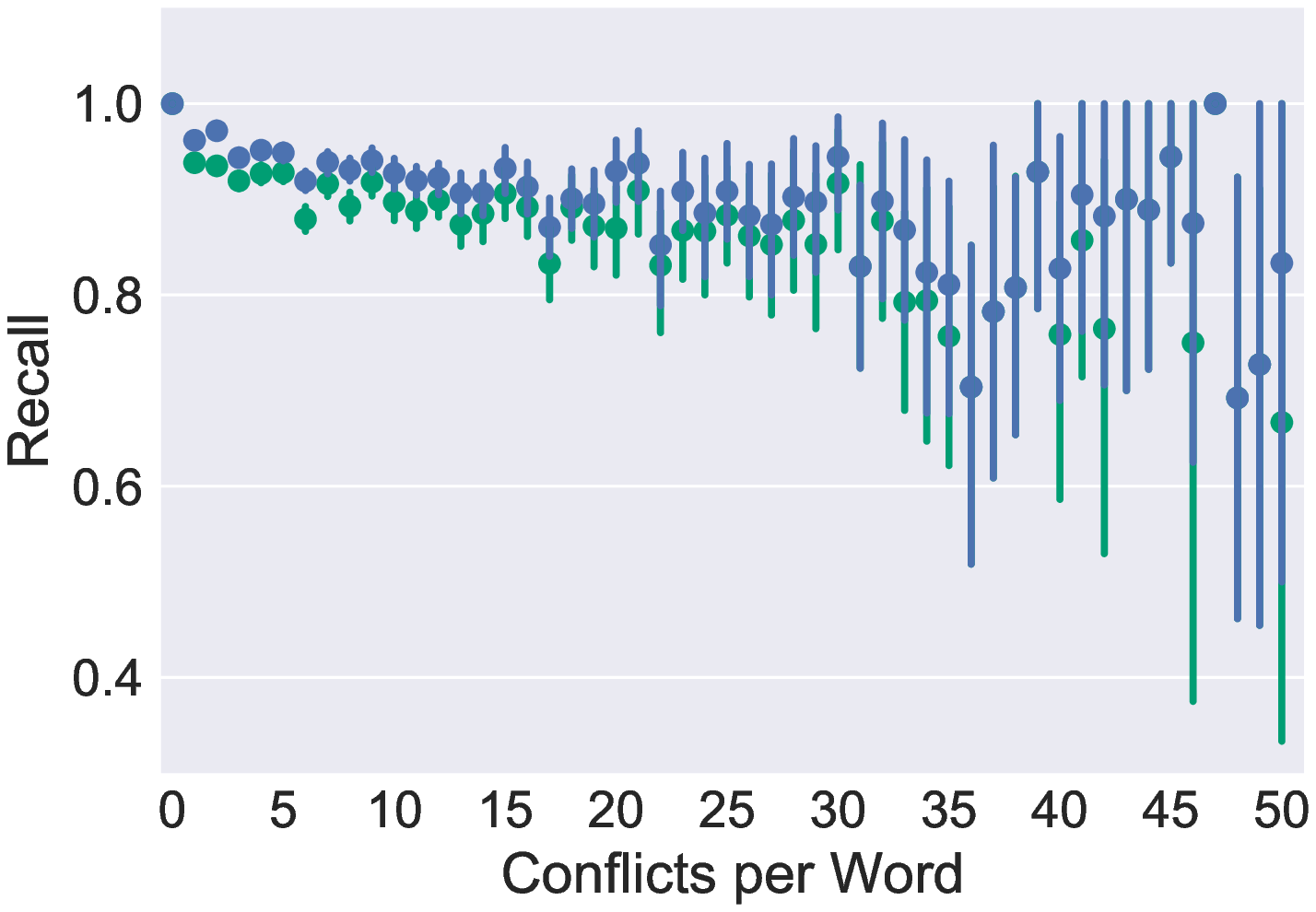}
  \vspace{-1.4em}

  \caption{Recall of ground truth nodes in input graph grouped on conflicts per node}

  \label{confl}
  \vspace{-3mm}
\end{figure}

\fi

\subsection{Fine Grained Analysis on Clique-EBM\footnote{For hyper-parameter settings, and other fine-grained analysis refer to \S 4 of the supplementary material.\label{expFo}}}

\paragraph{Pruning the edges in sentence graph $G$:} 
%Since Clique-EBM reports the best perofrmance in our experiments, we further analyze the performance of Clique-EBM by restricting the context window in which the nodes can be selected. That is, 
%We prune the edges from the sentence graph $G$ by considering the relative position of the words within the SHR output. 
%Our intuition to perform pruning of edges stems from the fact that 
Constructions in Sanskrit follow weak non-projectivity (with exception to verses), implying that they adhere to the principle of proximity~\cite{Ambakulkarni2015}. By proximity we expect that the words in a phrase go together, without being interrupted by a syntactically unrelated word. But the relative ordering between the phrases in a construction and the order of words within a phrase can be free. 

For any two words appearing in an exhaustive segmentation, we keep an edge only if both the words overlap within a distance of $k$ characters. We experiment with $k$ = 5, 10, 15 and 20. Hence, for $K=20$, a word will form edges with all the words that fall within 20 characters to left and 20 characters to right. The average length of an input sequence in DCS10K is 40.79 characters. We do not modify our inference procedure in system 8 other than to take care of the possibility that a clique need not always be returned.  Table \ref{window} shows the results for different values of $k$. Interestingly, the results show a monotonic increase with the increase in context window, and the results with the entire context are still better than those with $k=20$, even though only marginally. It is interesting to note that, keeping the entire context does not adversely affect the predictions as none of the pruned models outperforms System 8.

The lattice structure can be seen as an extreme case of pruning. %where an edge exists between only adjacent words in an exhaustive segmentation. 
We modify System 4 to use a non-sequential inference procedure, adapted from System 7. Here, the start and end markers were removed. Additionally, a given connected node pair has 2 edges, each in either of the directions.  We find that the model gives an F-Score of 87.4 which outperforms System 4 by more than three points.

\paragraph{Out of Vocabulary words:} 
As described in \S \ref{PCRW}, the distributional information from DCS is used as the corpus $\mathcal{C}$ for the feature generation. %It is possible that an inflected word form or a root in the input may not be present in $\mathcal{V}_w$ or $\mathcal{V}_r$ respectively. 
For the case of OOV in roots ($\mathcal{V}_r$), we use add-1 smoothing. But, for the case of OOV in inflections ($\mathcal{V}_w$) we find that using the evidence from corresponding root of the candidate is beneficial. %About 832 and 1236 of 1500 of the features use only the morphological and root information from the the source or the target node. 
DCS10k has 8,007 roots of which 514 are OOV and 833 occur only 1 to 5 times. The micro-averaged F-Score for these are 57.98 and 72.87, respectively. 

\renewcommand{\thetable}{\arabic{table}}
\begin{table}[]
\centering
\begin{small}
\begin{tabular}{|c|c|c|c|c|c|c|}
\hline
k & \multicolumn{3}{c|}{WPT} & \multicolumn{3}{c|}{WP3T} \\ \hline
 & P & R & F & P & R & F \\ \hline
5 & 90.46 & 92.27 & 91.36  &83.52  & 80.48  & 81.97  \\ \hline
10 & 92.92 & 95.07 & 93.98  & 85.32  & 84.4  & 84.86  \\ \hline
15 & 94.85 & 96.14 & 95.49   & 87.67  & 86.38  & 87.02 \\ \hline
20 & \textbf{95.23} & \textbf{96.49} &  \textbf{95.86} & \textbf{89.25}  & \textbf{88.62}  & \textbf{88.93}  \\ \hline
\end{tabular}
\end{small}
\caption{Performance of Clique-EBM with pruned edges in $G$. }
\label{window}

\end{table}

\if{}
\begin{table}[]
\centering
\begin{small}
\begin{tabular}{|l|l|l|}
\hline
\textbf{Type} & \textbf{Word Recall} & \textbf{Word++ Recall} \\ \hline
Noun & 96.869 & 88.998 \\ \hline
Verb & 95.911 & 94.424 \\ \hline
Compound & 93.518 & 91.067 \\ \hline
Indeclinable & 97.095 & 96.472 \\ \hline

\end{tabular}
\end{small}
\vspace{-2mm}
\caption{System performance on the coarse level POS}

\vspace{-4mm}
\end{table}
\fi

\begin{table}[]
\begin{small}
\begin{tabular}{cllll}
\hline
\multicolumn{1}{|c|}{\multirow{2}{*}{Type}} & \multicolumn{2}{c|}{WPT Recall}                                 & \multicolumn{2}{c|}{WP3T Recall}                                \\ \cline{2-5} 
\multicolumn{1}{|c|}{}                      & \multicolumn{1}{c|}{T-EBM} & \multicolumn{1}{c|}{C-EBM} & \multicolumn{1}{c|}{T-EBM} & \multicolumn{1}{c|}{C-EBM} \\ \hline
\multicolumn{1}{|c|}{Noun}                  & \multicolumn{1}{l|}{93.06}    & \multicolumn{1}{l|}{96.87}      & \multicolumn{1}{l|}{86.14}    & \multicolumn{1}{l|}{89.0}       \\ \hline
\multicolumn{1}{|c|}{Verb}                  & \multicolumn{1}{l|}{89.14}    & \multicolumn{1}{l|}{95.91}      & \multicolumn{1}{l|}{87.38}    & \multicolumn{1}{l|}{94.42}      \\ \hline
\multicolumn{1}{|c|}{Compound}              & \multicolumn{1}{l|}{89.35}    & \multicolumn{1}{l|}{93.52}      & \multicolumn{1}{l|}{86.01}    & \multicolumn{1}{l|}{91.07}      \\ \hline
\multicolumn{1}{|c|}{Indeclinable}          & \multicolumn{1}{l|}{95.07}    & \multicolumn{1}{l|}{97.09}      & \multicolumn{1}{l|}{94.93}    & \multicolumn{1}{l|}{96.47}      \\ \hline
                                            &                               &                                 &                               &                                
\end{tabular}
\end{small}
\caption{System performance on the coarse level POS for the competing systems Clique-EBM (C-EBM) and Tree-EBM (T-EBM)}
\label{morphWise}
\end{table}

\paragraph{Morphological class specific assessment}: Table \ref{morphWise} presents the micro-averaged recall for the words grouped based on their parts of speech (POS) for Clique-EBM and Tree-EBM. Both the systems follow similar trends and the morphological class misprediction is highest among the nouns and compounds (WP3T Recall). It also needs to be noted that the improvement made by Clique-EBM in comparison to Tree-EBM for WP3T was also on prediction of noun and compound morphological classes. Also in Tree-EBM, the mispredictions in compounds were mostly cases of the system getting the compound components confused to one of the morphological classes in nouns.

We find that considering the pairwise potential between all the words in a sentence in Clique-EBM led to improved morphological agreement between the words in comparison to Tree-EBM. In Tree-EBM, the top 5 cases of mispredictions from one morphological class to a particular wrong class were between those classes of nouns that differed in exactly one of the three possible grammatical categories, namely gender, number or declension, that makes up a noun. In Clique-EBM such patterns were not anymore present and more importantly the skewedness in such mispredictions  were considerably reduced.\footnote{Please refer to Tables 6 and 7 in the supplementary material for details}

Summarily, our non-sequential method of inference results in better performance in comparison to the sequential models. We also find that the sequential models see a drop in their performances when the number of words in a sentence increases. Leveraging the pairwise potentials between every connected nodes while making a prediction improves the performance. The performance gain of Clique-EBM over Tree-EBM illustrates the effectiveness of this approach. %We also find that restricting the neighbourhood of the candidate nodes, within a context window of 15 or 20 characters, leads to only a marginal decrease in the performance of the system. 

\section{Discussion}

%In sentence graph $G$, every candidate with a unique combination of morphemes (and position in the sentence) is considered a separate node,  %There is no distinction made between nodes that share morpheme and that do not. The edge formation is solely based on the conflicts in position of the candidates in a sentence.  the performance gain of clique-EBM over Tree-RBM illustrates 

 %The current sentence graph construction leads to a considerably large candidate space. But, this helped us to 

%The prediction of morphological tags in WP3T heavily gains from the clique based infernece of ourse. 
%The major gain due to our clique based inference procedure can be seen in the performance of our system in predicting the morphological classes (WP3T). 
In Sanskrit, syncretism \cite{crystal2011dictionary} %the morphological analysis of a word is often ambiguous as 
%a root can generate the same inflected word forms for different morphological tags, thereby making them homonyms of each other. This l
leads to ambiguity during morphological analysis. It can further be observed that such common root identical surface forms often have one or more common grammatical categories in their morphological classes \cite{goyal2016design}.  We find that the first three models in Table \ref{wppResults} often end up predicting an identical surface-form with an incorrect morphological tag, thus affecting the WP3T scores.\footnote{Details in \S4 of the Supplementary material}. The grammatical categories in a morphological class are indicative of the syntactic roles and the morphological agreement between the words in a construction. We empirically observe that the inference procedure for clique-EBM, which considers the entire input context and pairwise potentials between the candidates, helps in improving the performance of the model. A similar observation regarding incompatible morphological agreements between predicted words was made by \newcite{hassan2018achieving} for their NMT model. The authors introduced an elaborate 2 phase decoder and a  KL-Divergence based regularisation to combat the issue.

%In fact, for those 3 baseline systems, the top 5 morphological class mispredictions occur where the inflection is correct and the mispredicted morphologicla class share at least one grammatical category with that of the correct class

%In Sanskrit it is often the case that a root might genreate homonyms for inflections of differnt moprhological classes.  We find all the other 3 models as reported in Table \ref{wppResults} to the correct word with the same root. In fact . We find that for Clique-EBM this is not true. We couldnt find any specific patterns among the top misprediction of morpoholgoical classes in clique-EBM.  

%the models often ends up predicting homonyms with the same root, resulting in different  morphological classes. For all the three systems  

%results in a percentage gain of 11.64 in terms of F-Score from that of the next best system. We find that Tree-EBM and Lattice-EBM often end up  THis results in better segmentation but wider gap in predicting the morphological classes. In fact xx\% of the mis predictions for Tree-EBM is attributed morphological classes which differs only based on a some of the morphological classes. This genrrally happens with gendeer or number mismatch often indicating lapse in identifying morphologcal agreements between the words.

The energy based model (EBM) we propose is a general framework for structured prediction in Sanskrit. EBMs are widely used for various structured prediction tasks \cite{lecun2007energy,belanger2016structured}. \newcite{belanger2017deep} states that, ``CRFs are  typically  attributed  to \newcite{lafferty2001conditional},  but  many  of  the  core technical contributions of the paper appeared earlier in \newcite{lecun1998gradient}." GTNs \cite{lecun1998gradient}, in fact, work on a graph based input very similar to that of a lattice, a variant of which we use in L-EBM-Vanilla. 
For dependency parsing, use of a word-level lattice structure similar to \newcite{seeker2015graph}, where all the homonyms and syncertisms of a given surface-form form a lattice, will potentially result in a reduced candidate space than ours. Additionally, our model currently does not take into account the phrasal nature of compounds in Sanskrit \cite{lowe_2015}. This can further reduce the edge density in our current graph construction.  But, this needs further exploration, as current edge vectors may not be suitable for the task. To generate the possible candidates, we rely completely on the SHR. In case of words not recognised by the lexicon driven SHR, analysis of sentences with a partially recognised portion is still possible. Once a root is added, all its inflections can be generated by the SHR automatically.

\section{Conclusion}
\vspace{-2mm}

We proposed a novel approach to tackle word segmentation and morphological tagging problem in Sanskrit. %We extend the framework suggested by \newcite{krishna2016segmentation}. 
Our model outperforms \newcite{reddy2018seg}, the current state of the art, with an F-Score of 96.92. \newcite{reddy2018seg} report that the extension of their model to perform morphological tagging is not straightforward, as they learn a new sub-word vocabulary using the sentencepiece model \cite{schuster2012japanese}. %to alter the vocabulary so as to avoid sparsity of data. 

The free word order nature of the language motivated us to consider the input to be a graph so as to avoid the sequential processing of input. For the EBM we use, there is no requirement of proper normalisation \cite{lecun2006tutorial}. We benefit from this as we perform a search in the space of complete outputs and there is a combinatorial explosion in the output space for a linear increase in the input space \cite{doppa2014hc}.  The pre-training of the edge vectors with external knowledge in the form of morphological constraints %linguistic improves the performance of Chinese word segmentation systems. We find that capturing rich linguistic context through edge vectors 
is effective in reducing the task specific training size \cite{yang2017neural,andor-EtAl:2016:P16-1}. %In our case the external knowledge is captured by the PRA vectors. %This has helped in reducing the training data requirement more than 10 times what is needed for the current state of the art \cite{reddy2018seg}. 
%Our design decisions are primarily motivated from the fact that Sanskrit is a 
 % the entire input context needs to be considered while making the prediction. %We find that considering all the pairwise interactions leads to performance gain in the task. %The choice of our clique based inference procedure and loss function has . 
%Moreover,  %Thus, the use of energy based models makes the processing efficient even on large sentence graphs.  
%We decided to consider the candidate segments in a graph based formation so as to eliminate the sequential nature of the input and to enable our inference policy capture the entire input context.
%For example, changing the inference policy to search for a Steiner Tree \cite{voss1993worst,takahashi1980approximate}  instead of the clique reduce the word and word++ prediction task F-Score results to 90 \% and 82 \%. This is due to the lack of pairwise potentials available to the system. 
%But this inference policy is still  desirable for a model that needs to add dependency parsing, as performed in \newcite{seeker2015graph}, to the current system. Similarly, other structured prediction tasks such as arrangement of words in a sentence to its poetry order and prose order \cite{kulkarni2015} can be achieved using the Hamming path algorithm on a complete graph algorithm as our inference procedure. Though this essentially is a permutation or a sorting problem, since we have directed edges, there is only one unique way to construct a connected path between given set of edges.

\section*{Acknowledgements}

We are grateful to Oliver Hellwig for providing the
DCS Corpus and G\'erard Huet for providing the Sanskrit Heritage Engine, to be installed at local systems. We extend our gratitude to Amba Kulkarni and Rogers Mathew, along with G\'erard for helpful comments and discussions regarding the work. We thank the anonymous reviewers for their constructive and helpful comments, which greatly improved the paper. We are indebted to Unni Krishnan T A for his contributions towards implementation of the framework.

\if{}
\section{Credits}

This document has been adapted from the instructions for earlier ACL
and NAACL proceedings. It represents a recent build from
\url{https://github.com/acl-org/acl-pub}, with modifications by Micha
Elsner and Preethi Raghavan, based on the NAACL 2018 instructions by
Margaret Michell and Stephanie Lukin, 2017/2018 (NA)ACL bibtex
suggestions from Jason Eisner, ACL 2017 by Dan Gildea and Min-Yen Kan,
NAACL 2017 by Margaret Mitchell, ACL 2012 by Maggie Li and Michael
White, those from ACL 2010 by Jing-Shing Chang and Philipp Koehn,
those for ACL 2008 by Johanna D. Moore, Simone Teufel, James Allan,
and Sadaoki Furui, those for ACL 2005 by Hwee Tou Ng and Kemal
Oflazer, those for ACL 2002 by Eugene Charniak and Dekang Lin, and
earlier ACL and EACL formats.  Those versions were written by several
people, including John Chen, Henry S. Thompson and Donald
Walker. Additional elements were taken from the formatting
instructions of the {\em International Joint Conference on Artificial
  Intelligence} and the \emph{Conference on Computer Vision and
  Pattern Recognition}.

\section{Introduction}

The following instructions are directed to authors of papers submitted
to \confname{} or accepted for publication in its proceedings. All
authors are required to adhere to these specifications. Authors are
required to provide a Portable Document Format (PDF) version of their
papers. \textbf{The proceedings are designed for printing on A4
paper.}

\section{General Instructions}

Manuscripts must be in two-column format.  Exceptions to the
two-column format include the title, authors' names and complete
addresses, which must be centered at the top of the first page, and
any full-width figures or tables (see the guidelines in
Subsection~\ref{ssec:first}). {\bf Type single-spaced.}  Start all
pages directly under the top margin. See the guidelines later
regarding formatting the first page.  The manuscript should be
printed single-sided and its length
should not exceed the maximum page limit described in Section~\ref{sec:length}.
Pages are numbered for  initial submission. However, {\bf do not number the pages in the camera-ready version}.

By uncommenting {\small\verb|\aclfinalcopy|} at the top of this 
 document, it will compile to produce an example of the camera-ready formatting; by leaving it commented out, the document will be anonymized for initial submission.  When you first create your submission on softconf, please fill in your submitted paper ID where {\small\verb|***|} appears in the {\small\verb|\def\aclpaperid{***}|} definition at the top.

The review process is double-blind, so do not include any author information (names, addresses) when submitting a paper for review.  
However, you should maintain space for names and addresses so that they will fit in the final (accepted) version.  The \confname{} \LaTeX\ style will create a titlebox space of 2.5in for you when {\small\verb|\aclfinalcopy|} is commented out.  

The author list for submissions should include all (and only) individuals who made substantial contributions to the work presented. Each author listed on a submission to \confname{} will be notified of submissions, revisions and the final decision. No authors may be added to or removed from submissions to \confname{} after the submission deadline.

\subsection{The Ruler}
The \confname{} style defines a printed ruler which should be presented in the
version submitted for review.  The ruler is provided in order that
reviewers may comment on particular lines in the paper without
circumlocution.  If you are preparing a document without the provided
style files, please arrange for an equivalent ruler to
appear on the final output pages.  The presence or absence of the ruler
should not change the appearance of any other content on the page.  The
camera ready copy should not contain a ruler. (\LaTeX\ users may uncomment the {\small\verb|\aclfinalcopy|} command in the document preamble.)  

Reviewers: note that the ruler measurements do not align well with
lines in the paper -- this turns out to be very difficult to do well
when the paper contains many figures and equations, and, when done,
looks ugly. In most cases one would expect that the approximate
location will be adequate, although you can also use fractional
references ({\em e.g.}, the first paragraph on this page ends at mark $108.5$).

\subsection{Electronically-available resources}

\conforg{} provides this description in \LaTeX2e{} ({\small\tt
  emnlp2018.tex}) and PDF format ({\small\tt emnlp2018.pdf}), along
with the \LaTeX2e{} style file used to format it ({\small\tt
  emnlp2018.sty}) and an ACL bibliography style ({\small\tt
  acl\_natbib\_nourl.bst}) and example bibliography ({\small\tt
  emnlp2018.bib}).  These files are all available at
\url{http://emnlp2018.org/downloads/emnlp18-latex.zip}; a Microsoft
Word template file ({\small\tt emnlp18-word.docx}) and example
submission pdf ({\small\tt emnlp18-word.pdf}) is available at
\url{http://emnlp2018.org/downloads/emnlp18-word.zip}.  We strongly
recommend the use of these style files, which have been appropriately
tailored for the \confname{} proceedings.

\subsection{Format of Electronic Manuscript}
\label{sect:pdf}

For the production of the electronic manuscript you must use Adobe's
Portable Document Format (PDF). PDF files are usually produced from
\LaTeX\ using the \textit{pdflatex} command. If your version of
\LaTeX\ produces Postscript files, you can convert these into PDF
using \textit{ps2pdf} or \textit{dvipdf}. On Windows, you can also use
Adobe Distiller to generate PDF.

Please make sure that your PDF file includes all the necessary fonts
(especially tree diagrams, symbols, and fonts with Asian
characters). When you print or create the PDF file, there is usually
an option in your printer setup to include none, all or just
non-standard fonts.  Please make sure that you select the option of
including ALL the fonts. \textbf{Before sending it, test your PDF by
  printing it from a computer different from the one where it was
  created.} Moreover, some word processors may generate very large PDF
files, where each page is rendered as an image. Such images may
reproduce poorly. In this case, try alternative ways to obtain the
PDF. One way on some systems is to install a driver for a postscript
printer, send your document to the printer specifying ``Output to a
file'', then convert the file to PDF.

It is of utmost importance to specify the \textbf{A4 format} (21 cm
x 29.7 cm) when formatting the paper. When working with
{\tt dvips}, for instance, one should specify {\tt -t a4}.
Or using the command \verb|\special{papersize=210mm,297mm}| in the latex
preamble (directly below the \verb|\usepackage| commands). Then using 
{\tt dvipdf} and/or {\tt pdflatex} which would make it easier for some.

Print-outs of the PDF file on A4 paper should be identical to the
hardcopy version. If you cannot meet the above requirements about the
production of your electronic submission, please contact the
publication chairs as soon as possible.

\subsection{Layout}
\label{ssec:layout}

Format manuscripts two columns to a page, in the manner these
instructions are formatted. The exact dimensions for a page on A4
paper are:

\begin{itemize}
\item Left and right margins: 2.5 cm
\item Top margin: 2.5 cm
\item Bottom margin: 2.5 cm
\item Column width: 7.7 cm
\item Column height: 24.7 cm
\item Gap between columns: 0.6 cm
\end{itemize}

\noindent Papers should not be submitted on any other paper size.
 If you cannot meet the above requirements about the production of 
 your electronic submission, please contact the publication chairs 
 above as soon as possible.

\subsection{Fonts}

For reasons of uniformity, Adobe's {\bf Times Roman} font should be
used. In \LaTeX2e{} this is accomplished by putting

\begin{quote}
\begin{verbatim}
\usepackage{times}
\usepackage{latexsym}
\end{verbatim}
\end{quote}
in the preamble. If Times Roman is unavailable, use {\bf Computer
  Modern Roman} (\LaTeX2e{}'s default).  Note that the latter is about
  10\% less dense than Adobe's Times Roman font.

\begin{table}[t!]
\begin{center}
\begin{tabular}{|l|rl|}
\hline \bf Type of Text & \bf Font Size & \bf Style \\ \hline
paper title & 15 pt & bold \\
author names & 12 pt & bold \\
author affiliation & 12 pt & \\
the word ``Abstract'' & 12 pt & bold \\
section titles & 12 pt & bold \\
document text & 11 pt  &\\
captions & 10 pt & \\
abstract text & 10 pt & \\
bibliography & 10 pt & \\
footnotes & 9 pt & \\
\hline
\end{tabular}
\end{center}
\caption{\label{font-table} Font guide. }
\end{table}

\subsection{The First Page}
\label{ssec:first}

Center the title, author's name(s) and affiliation(s) across both
columns. Do not use footnotes for affiliations. Do not include the
paper ID number assigned during the submission process. Use the
two-column format only when you begin the abstract.

{\bf Title}: Place the title centered at the top of the first page, in
a 15-point bold font. (For a complete guide to font sizes and styles,
see Table~\ref{font-table}) Long titles should be typed on two lines
without a blank line intervening. Approximately, put the title at 2.5
cm from the top of the page, followed by a blank line, then the
author's names(s), and the affiliation on the following line. Do not
use only initials for given names (middle initials are allowed). Do
not format surnames in all capitals ({\em e.g.}, use ``Mitchell'' not
``MITCHELL'').  Do not format title and section headings in all
capitals as well except for proper names (such as ``BLEU'') that are
conventionally in all capitals.  The affiliation should contain the
author's complete address, and if possible, an electronic mail
address. Start the body of the first page 7.5 cm from the top of the
page.

The title, author names and addresses should be completely identical
to those entered to the electronical paper submission website in order
to maintain the consistency of author information among all
publications of the conference. If they are different, the publication
chairs may resolve the difference without consulting with you; so it
is in your own interest to double-check that the information is
consistent.

{\bf Abstract}: Type the abstract at the beginning of the first
column. The width of the abstract text should be smaller than the
width of the columns for the text in the body of the paper by about
0.6 cm on each side. Center the word {\bf Abstract} in a 12 point bold
font above the body of the abstract. The abstract should be a concise
summary of the general thesis and conclusions of the paper. It should
be no longer than 200 words. The abstract text should be in 10 point font.

{\bf Text}: Begin typing the main body of the text immediately after
the abstract, observing the two-column format as shown in

the present document. Do not include page numbers.

{\bf Indent}: Indent when starting a new paragraph, about 0.4 cm. Use 11 points for text and subsection headings, 12 points for section headings and 15 points for the title.

\begin{table}
\centering
\small
\begin{tabular}{cc}
\begin{tabular}{|l|l|}
\hline
{\bf Command} & {\bf Output}\\\hline
\verb|{\"a}| & {\"a} \\
\verb|{\^e}| & {\^e} \\
\verb|{\`i}| & {\`i} \\ 
\verb|{\.I}| & {\.I} \\ 
\verb|{\o}| & {\o} \\
\verb|{\'u}| & {\'u}  \\ 
\verb|{\aa}| & {\aa}  \\\hline
\end{tabular} & 
\begin{tabular}{|l|l|}
\hline
{\bf Command} & {\bf  Output}\\\hline
\verb|{\c c}| & {\c c} \\ 
\verb|{\u g}| & {\u g} \\ 
\verb|{\l}| & {\l} \\ 
\verb|{\~n}| & {\~n} \\ 
\verb|{\H o}| & {\H o} \\ 
\verb|{\v r}| & {\v r} \\ 
\verb|{\ss}| & {\ss} \\\hline
\end{tabular}
\end{tabular}
\caption{Example commands for accented characters, to be used in, {\em e.g.}, \BibTeX\ names.}\label{tab:accents}
\end{table}

\subsection{Sections}

{\bf Headings}: Type and label section and subsection headings in the
style shown on the present document.  Use numbered sections (Arabic
numerals) in order to facilitate cross references. Number subsections
with the section number and the subsection number separated by a dot,
in Arabic numerals.
Do not number subsubsections.

\begin{table*}[t!]
\centering
\begin{tabular}{lll}
  output & natbib & previous \conforg{} style files\\
  \hline
  \citep{Gusfield:97} & \verb|\citep| & \verb|\cite| \\
  \citet{Gusfield:97} & \verb|\citet| & \verb|\newcite| \\
  \citeyearpar{Gusfield:97} & \verb|\citeyearpar| & \verb|\shortcite| \\
\end{tabular}
\caption{Citation commands supported by the style file.
  The citation style is based on the natbib package and
  supports all natbib citation commands.
  It also supports commands defined in previous \conforg{} style files
  for compatibility.
  }
\end{table*}

{\bf Citations}: Citations within the text appear in parentheses
as~\cite{Gusfield:97} or, if the author's name appears in the text
itself, as Gusfield~\shortcite{Gusfield:97}.
Using the provided \LaTeX\ style, the former is accomplished using
{\small\verb|\cite|} and the latter with {\small\verb|\shortcite|} or {\small\verb|\newcite|}. Collapse multiple citations as in~\cite{Gusfield:97,Aho:72}; this is accomplished with the provided style using commas within the {\small\verb|\cite|} command, {\em e.g.}, {\small\verb|\cite{Gusfield:97,Aho:72}|}. Append lowercase letters to the year in cases of ambiguities.  
 Treat double authors as
in~\cite{Aho:72}, but write as in~\cite{Chandra:81} when more than two
authors are involved. Collapse multiple citations as
in~\cite{Gusfield:97,Aho:72}. Also refrain from using full citations
as sentence constituents.

We suggest that instead of
\begin{quote}
  ``\cite{Gusfield:97} showed that ...''
\end{quote}
you use
\begin{quote}
``Gusfield \shortcite{Gusfield:97}   showed that ...''
\end{quote}

If you are using the provided \LaTeX{} and Bib\TeX{} style files, you
can use the command \verb|\citet| (cite in text)
to get ``author (year)'' citations.

If the Bib\TeX{} file contains DOI fields, the paper
title in the references section will appear as a hyperlink
to the DOI, using the hyperref \LaTeX{} package.
To disable the hyperref package, load the style file
with the \verb|nohyperref| option: \\{\small
\verb|\usepackage[nohyperref]{acl2018}|}

\textbf{Digital Object Identifiers}: As part of our work to make ACL
materials more widely used and cited outside of our discipline, ACL
has registered as a CrossRef member, as a registrant of Digital Object
Identifiers (DOIs), the standard for registering permanent URNs for
referencing scholarly materials. \conforg{} has \textbf{not} adopted the
ACL policy of requiring camera-ready references to contain the appropriate
  DOIs (or as a second resort, the hyperlinked ACL Anthology
  Identifier). But we certainly encourage you to use
  Bib\TeX\ records that contain DOI or URLs for any of the ACL
  materials that you reference. Appropriate records should be found
for most materials in the current ACL Anthology at
\url{http://aclanthology.info/}.

As examples, we cite \cite{P16-1001} to show you how papers with a DOI
will appear in the bibliography.  We cite \cite{C14-1001} to show how
papers without a DOI but with an ACL Anthology Identifier will appear
in the bibliography.  

\textbf{Anonymity:} As reviewing will be double-blind, the submitted
version of the papers should not include the authors' names and
affiliations. Furthermore, self-references that reveal the author's
identity, {\em e.g.},
\begin{quote}
``We previously showed \cite{Gusfield:97} ...''  
\end{quote}
should be avoided. Instead, use citations such as 
\begin{quote}
``\citeauthor{Gusfield:97} \shortcite{Gusfield:97}
previously showed ... ''
\end{quote}

Preprint servers such as arXiv.org and workshops that do not
have published proceedings are not considered archival for purposes of
submission. However, to preserve the spirit of blind review, authors
are encouraged to refrain from posting until the completion of the
review process. Otherwise, authors must state in the online submission
form the name of the workshop or preprint server and title of the
non-archival version. The submitted version should be suitably
anonymized and not contain references to the prior non-archival
version. Reviewers will be told: ``The author(s) have notified us that
there exists a non-archival previous version of this paper with
significantly overlapping text. We have approved submission under
these circumstances, but to preserve the spirit of blind review, the
current submission does not reference the non-archival version.''

\textbf{Please do not use anonymous citations} and do not include
 when submitting your papers. Papers that do not
conform to these requirements may be rejected without review.

\textbf{References}: Gather the full set of references together under
the heading {\bf References}; place the section before any Appendices,
unless they contain references. Arrange the references alphabetically
by first author, rather than by order of occurrence in the text.
By using a .bib file, as in this template, this will be automatically 
handled for you. See the \verb|\bibliography| commands near the end for more.

Provide as complete a citation as possible, using a consistent format,
such as the one for {\em Computational Linguistics\/} or the one in the 
{\em Publication Manual of the American 
Psychological Association\/}~\cite{APA:83}. Use of full names for
authors rather than initials is preferred. A list of abbreviations
for common computer science journals can be found in the ACM 
{\em Computing Reviews\/}~\cite{ACM:83}.

The \LaTeX{} and Bib\TeX{} style files provided roughly fit the
American Psychological Association format, allowing regular citations, 
short citations and multiple citations as described above.  

\begin{itemize}
\item Example citing an arxiv paper: \cite{rasooli-tetrault-2015}. 
\item Example article in journal citation: \cite{Ando2005}.
\item Example article in proceedings, with location: \cite{borsch2011}.
\item Example article in proceedings, without location: \cite{andrew2007scalable}.
\end{itemize}
See corresponding .bib file for further details.

Submissions should accurately reference prior and related work, including code and data. If a piece of prior work appeared in multiple venues, the version that appeared in a refereed, archival venue should be referenced. If multiple versions of a piece of prior work exist, the one used by the authors should be referenced. Authors should not rely on automated citation indices to provide accurate references for prior and related work.

{\bf Appendices}: Appendices, if any, directly follow the text and the
references (but see above).  Letter them in sequence and provide an
informative title: {\bf Appendix A. Title of Appendix}.

\subsection{URLs}

URLs can be typeset using the \verb|\url| command. However, very long
URLs cause a known issue in which the URL highlighting may incorrectly
cross pages or columns in the document. Please check carefully for
URLs too long to appear in the column, which we recommend you break,
shorten or place in footnotes. Be aware that actual URL should appear
in the text in human-readable format; neither internal nor external
hyperlinks will appear in the proceedings.

\subsection{Footnotes}

{\bf Footnotes}: Put footnotes at the bottom of the page and use 9
point font. They may be numbered or referred to by asterisks or other
symbols.\footnote{This is how a footnote should appear.} Footnotes
should be separated from the text by a line.\footnote{Note the line
separating the footnotes from the text.}

\subsection{Graphics}

{\bf Illustrations}: Place figures, tables, and photographs in the
paper near where they are first discussed, rather than at the end, if
possible.  Wide illustrations may run across both columns.  Color
illustrations are discouraged, unless you have verified that  
they will be understandable when printed in black ink.

{\bf Captions}: Provide a caption for every illustration; number each one
sequentially in the form:  ``Figure 1. Caption of the Figure.'' ``Table 1.
Caption of the Table.''  Type the captions of the figures and 
tables below the body, using 11 point text.

\subsection{Accessibility}
\label{ssec:accessibility}

In an effort to accommodate people who are color-blind (as well as those printing
to paper), grayscale readability for all accepted papers will be
encouraged.  Color is not forbidden, but authors should ensure that
tables and figures do not rely solely on color to convey critical
distinctions. A simple criterion: All curves and points in your figures should be clearly distinguishable without color.

% Min: no longer used as of ACL 2018, following ACL exec's decision to
% remove this extra workflow that was not executed much.
% BEGIN: remove
%% \section{XML conversion and supported \LaTeX\ packages}

%% Following ACL 2014 we will also we will attempt to automatically convert 
%% your \LaTeX\ source files to publish papers in machine-readable 
%% XML with semantic markup in the ACL Anthology, in addition to the 
%% traditional PDF format.  This will allow us to create, over the next 
%% few years, a growing corpus of scientific text for our own future research, 
%% and picks up on recent initiatives on converting ACL papers from earlier 
%% years to XML. 

%% We encourage you to submit a ZIP file of your \LaTeX\ sources along
%% with the camera-ready version of your paper. We will then convert them
%% to XML automatically, using the LaTeXML tool
%% (\url{http://dlmf.nist.gov/LaTeXML}). LaTeXML has \emph{bindings} for
%% a number of \LaTeX\ packages, including the ACL 2018 stylefile. These
%% bindings allow LaTeXML to render the commands from these packages
%% correctly in XML. For best results, we encourage you to use the
%% packages that are officially supported by LaTeXML, listed at
%% \url{http://dlmf.nist.gov/LaTeXML/manual/included.bindings}
% END: remove

\section{Translation of non-English Terms}

It is also advised to supplement non-English characters and terms
with appropriate transliterations and/or translations
since not all readers understand all such characters and terms.
Inline transliteration or translation can be represented in
the order of: original-form transliteration ``translation''.

\section{Length of Submission}
\label{sec:length}

The \confname{} main conference accepts submissions of long papers and
short papers.
 Long papers may consist of up to eight (8) pages of
content plus unlimited pages for references. Upon acceptance, final
versions of long papers will be given one additional page -- up to nine (9)
pages of content plus unlimited pages for references -- so that reviewers' comments
can be taken into account. Short papers may consist of up to four (4)
pages of content, plus unlimited pages for references. Upon
acceptance, short papers will be given five (5) pages in the
proceedings and unlimited pages for references. 

For both long and short papers, all illustrations and tables that are part
of the main text must be accommodated within these page limits, observing
the formatting instructions given in the present document. Supplementary
material in the form of appendices does not count towards the page limit; see appendix A for further information.

However, note that supplementary material should be supplementary
(rather than central) to the paper, and that reviewers may ignore
supplementary material when reviewing the paper (see Appendix
\ref{sec:supplemental}). Papers that do not conform to the specified
length and formatting requirements are subject to be rejected without
review.

Workshop chairs may have different rules for allowed length and
whether supplemental material is welcome. As always, the respective
call for papers is the authoritative source.

\section*{Acknowledgments}

The acknowledgments should go immediately before the references.  Do
not number the acknowledgments section. Do not include this section
when submitting your paper for review. \\

\noindent {\bf Preparing References:} \\

Include your own bib file like this:

{\small\verb|\bibliographystyle{acl_natbib_nourl}|
\verb|\bibliography{emnlp2018}|}

Where \verb|emnlp2018| corresponds to the {\tt emnlp2018.bib} file.
\fi

\bibliography{emnlp2018}
\bibliographystyle{acl_natbib_nourl}

\if{}

\appendix
\section{Supplemental Material}
\label{sec:supplemental}
Each \confname{} submission can be accompanied by a single PDF
appendix, one {\small\tt.tgz} or {\small\tt.zip} appendix containing
software, and one {\small\tt.tgz} or {\small\tt.zip} appendix
containing data.

Submissions may include resources (software and/or data) used in in
the work and described in the paper. Papers that are submitted with
accompanying software and/or data may receive additional credit toward
the overall evaluation score, and the potential impact of the software
and data will be taken into account when making the
acceptance/rejection decisions. Any accompanying software and/or data
should include licenses and documentation of research review as
appropriate.

\confname{} also encourages the submission of supplementary material
to report preprocessing decisions, model parameters, and other details
necessary for the replication of the experiments reported in the
paper. Seemingly small preprocessing decisions can sometimes make a
large difference in performance, so it is crucial to record such
decisions to precisely characterize state-of-the-art methods.

Nonetheless, supplementary material should be supplementary (rather
than central) to the paper. {\bf Submissions that misuse the supplementary 
material may be rejected without review.}
Essentially, supplementary material may include explanations or details
of proofs or derivations that do not fit into the paper, lists of
features or feature templates, sample inputs and outputs for a system,
pseudo-code or source code, and data. (Source code and data should
be separate uploads, rather than part of the paper).

The paper should not rely on the supplementary material: while the paper
may refer to and cite the supplementary material and the supplementary material will be available to the
reviewers, they will not be asked to review the
supplementary material.

Appendices ({\em i.e.} supplementary material in the form of proofs, tables,
or pseudo-code) should be {\bf uploaded as supplementary material} when submitting the paper for review.
Upon acceptance, the appendices come after the references, as shown here. Use
\verb|\appendix| before any appendix section to switch the section
numbering over to letters.
\fi
\end{document}